\documentclass{article}

\usepackage{arxiv}

\usepackage[utf8]{inputenc} 
\usepackage[T1]{fontenc}    
\usepackage{hyperref}       
\usepackage{url}            
\usepackage{booktabs}       
\usepackage{amsfonts}       
\usepackage{amssymb}
\usepackage{amsmath}
\usepackage[ruled,vlined]{algorithm2e}
\usepackage{multirow}
\usepackage{nicefrac}       
\usepackage{microtype}      
\usepackage{lipsum}
\usepackage{graphicx}
\usepackage{xcolor}
\usepackage{pgfplots}
\usepgfplotslibrary{groupplots}
\graphicspath{ {./images/} }
\pgfplotsset{compat=1.18}

\definecolor{betterfill}{RGB}{210,230,210}
\definecolor{samefill}{RGB}{240,240,240}
\definecolor{worsefill}{RGB}{235,210,210}

\title{On the Use of Iterative Problem Solving for the Traveling Salesperson Problem with Changing Time Window Constraints}

\author{
 Hy Nguyen, Thanh Nguyen Pham, Helen Yuliana Angmalisang, Liam Wigney, Frank Neumann\\
 Optimisation and Logistics\\
 School of Computer Science and Information Technology\\
 Adelaide University\\
 Adelaide, Australia\\
 \texttt{hy.nguyen@student.adelaide.edu.au, thanhnguyen.pham@student.adelaide.edu.au}\\
 \texttt{helen.angmalisang@adelaide.edu.au, liam.wigney@adelaide.edu.au}\\
 \texttt{frank.neumann@adelaide.edu.au}
}

\begin{document}
\maketitle

\begin{abstract}
In many real-world settings, problem instances that need to be solved are quite similar, and knowledge from previous optimization runs can potentially be utilized. We explore this for the Traveling Salesperson problem with time windows (TSPTW), which often arises in settings where the travel-time matrix is fixed but time-window constraints change across related tasks. Existing TSPTW studies, however, have not systematically compared solving such task sequences independently with sequential transfer from previously solved tasks. We address this gap using a multi-task benchmark in which each base instance is expanded into five related tasks under two environments: partial time-window expansion and swap-additive time reassignment. We compare a standard from-scratch protocol with an iterative protocol that initializes each task from the best tour of the previous task, using the popular local search approaches LNS, VNS, and LKH-3 under a common penalized-score objective.
Our experimental results show that the iterative protocol is consistently superior in the progressive-relaxation setting and generally competitive under swap-additive changes, with improvements increasing on more difficult instances.

\end{abstract}

\section{Introduction}

Heuristic search algorithms, such as local search, simulated annealing, and evolutionary algorithms, have been applied to a wide range of combinatorial optimization problems \cite{NummanAliAbbas2026Nhtf,PanwarKaruna2026Caoe,RoostapourVahid2022Same}. One of the famous combinatorial optimization problems is the Traveling Salesperson Problem (TSP). In numerous studies, TSP can involve additional constraints \cite{GouveiaLuis2025Amft,PanwarKaruna2026Caoe}, including time windows \cite{DoppstadtChristian2020THEV} that require the arrival of the salesperson within given time intervals. The TSP with Time Windows (TSPTW) has been studied for decades \cite{DumasTSPTW1995,LOPEZIBANEZ20133806} with many heuristic algorithms available to solve it. Some of them are the Large Neighborhood Search (LNS) \cite{Shaw1998UsingCP}, the Variable Neighborhood Search (VNS) \cite{DaSilvaUrrutia2010}, and the Lin-Kernighan-Helsgaun (LKH-3) algorithm \cite{helsgaun2017extension}.

In many real-world settings, problems only slightly change over time. In the service provider industry, for example, there may be additional treatments only for a small number of patients on some occasions or slight changes in maintenance schedules for only some facilities, which form a series of similar optimization problems \cite{GschwindTimo2015EHoD}. Meanwhile, solutions obtained for earlier problems in a series may help compute good solutions for later related problems \cite{ZhengXiaolong2020SEMO}. This is the basic idea behind transfer optimization \cite{Gupta2018}. It takes advantage of the transfer of knowledge between related problems.

It has been shown in~\cite{LimRay2019Crpb} that knowledge transfer from past experience can benefit route planning. For TSP without time windows, solution transfer for the TSP using matching algorithms has been investigated in~\cite{Wigney2025Transfer}. In the Vehicle Routing Problem (VRP), which is a generalization of TSP, the transfer optimization framework performs better than the state-of-the-art VRP solvers~\cite{GuoVRP2025}. In contrast, it has been stated in~\cite{XuHao2022EMOW} that transferring solutions between tasks may cause difficulties rather than helping the algorithms to find the optimal solution.

Sequential transfer is one category of transfer optimization~\cite{Gupta2018}, where there is transfer of knowledge between problems. In sequential transfer, each task is solved one-by-one with knowledge transfer between each task. With the transferred knowledge, the computation cost may be cheaper compared to the classical method that solves each instance independently.

Related to this is work on dynamic problems where problem formulations frequently change over time. Investigations have been carried out where such changes impact the objective function as well as constraints of the considered problems \cite{DBLP:journals/tcyb/ZhangMNZ21,DBLP:journals/tec/WangXJT25,DBLP:journals/tec/QiaoYQ0SYLT23}. Investigations here focus on a relatively high frequency of changes, whereas for iterative problem solving, we allow each algorithm on each instance a relatively large computational budget.

To the best of our knowledge, no prior work has systematically compared from-scratch and sequential protocols for related TSPTW task sequences. Therefore, in this study, we examine how sequential transfer affects the optimization process when solving TSPTW instances. We are especially interested in the time-window aspect of the problem. Hence, our settings consider a fixed TSP instance together with different time window constraints. $k$ problem instances are given sequentially, and we investigate how previously solved problem instances are helpful in solving the remaining ones.

\subsection{Our Contribution}

We investigate how TSPTW instances with different time window constraints can be solved by local search methods.
The setup is motivated by scenarios in which the TSP distance matrix is fixed, but the time window requirement changes from day to day.
We compare classical local search approaches for the TSPTW with iterative variants that reuse solutions from previously solved instances, and examine whether this improves feasibility and final solution quality under a fixed function-evaluation budget.

For our experimental investigations, we generate a set of benchmark instances for multi-task TSPTW. Using the instances, the performance of several TSP heuristics, namely the Large Neighborhood Search (LNS), Variable Neighborhood Search (VNS), and Lin-Kernighan-Helsgaun (LKH-3) algorithms are evaluated under different solving protocols. We investigate how sequentially transferring the solutions between tasks benefits the optimization process. We compare sequential transfer against the standard approach of solving each task independently. Our experiments show that sequential transfer can substantially improve performance on TSPTW when consecutive tasks share a similar time-window structure. In settings where time windows change gradually, the iterative variants of LNS, VNS, and LKH-3 often achieved higher feasibility rates and better solution quality than solving each instance from scratch under the same function-evaluation budget. When the changes between tasks were more disruptive, the benefits became smaller, but the iterative approach remained competitive overall.

The remainder of this paper is structured as follows. Section~\ref{sec:tsptw} introduces the TSPTW setting and the sequential transfer framework considered in this work. Section~\ref{sec:heuristics} describes the heuristic algorithms used in our study, including LNS, VNS, and LKH-3, together with their iterative variants. Section~\ref{sec:experiments} presents the benchmark generation process and the experimental setup. Section~\ref{sec:results} reports and discusses the experimental results, highlighting when and how sequential transfer improves performance. Finally, Section~\ref{sec:conclusion} concludes the paper and outlines directions for future research.

\section{Traveling Salesperson Problem with Time Windows (TSPTW)}
\label{sec:tsptw}

We use the standard notation for the Traveling Salesperson Problem with Time Windows (TSPTW) as defined by \cite{DumasTSPTW1995}, where travel costs are given by an edge-cost (travel-time) function. An instance consists of a set of cities $V=\{v_1,\ldots,v_n\}$ and a travel-time matrix $d: V \times V \rightarrow \mathbb{R}_{\ge 0}$, where $d(u,v)$ denotes the travel time from city $u$ to city $v$. In addition, each city $v_i$ is associated with a time window $[a_i,b_i]$, specifying the allowable interval for visiting city $v_i$. Arriving early is permitted, and waiting is allowed.

A tour is represented by a permutation $\pi=(\pi_1,\ldots,\pi_n)$ of the cities in $V$. The tour cost is defined as
\[
c(\pi) = d(\pi_n,\pi_1) + \sum_{i=1}^{n-1} d(\pi_i,\pi_{i+1}).
\]

To evaluate feasibility with respect to time windows, we define the induced schedule of a tour recursively. Let $\tau_i(\pi)$ denote the arrival time at city $\pi_i$ and $s_i(\pi)$ the service start time after waiting. We set $\tau_1(\pi)=0$ and for $i=1,\ldots,n$ define
\[
s_i(\pi)=\max\{\tau_i(\pi), a_{\pi_i}\},
\qquad
\tau_{i+1}(\pi)=s_i(\pi)+d(\pi_i,\pi_{i+1}),
\]
where $\pi_{n+1}=\pi_1$ closes the tour. The time-window constraint violation is defined as the total lateness over all visited cities:
\[
\mathrm{CV}(\pi)=\sum_{i=1}^{n}\max\{0, s_i(\pi)-b_{\pi_i}\}.
\]
A tour is feasible if and only if $\mathrm{CV}(\pi)=0$.

We evaluate solutions using the penalized score
\[
\mathrm{score}(\pi) = c(\pi) + \mathrm{CV}(\pi)\cdot L,
\qquad
L = \sum_{u\in V}\sum_{v\in V} d(u,v).
\]
Here, $L$ is a fixed constant computed once per instance from the travel-time matrix. The setting implies that each unit of constraint violation with respect to $\mathrm{CV}(\pi)$ outweighs the cost of any tour. This objective allows meaningful comparison even when feasibility is not reached within the allocated budget, as it focuses on the amount of constraint violation.

\section{Algorithms and Parameters}
\label{sec:heuristics}

To evaluate the different protocols in this problem, we investigate popular local search algorithms, namely LNS, VNS, and LKH-3 for TSPTW. In the following, we give a brief description of these prominent local search approaches available in the literature.

\subsection{Large Neighborhood Search (LNS)}

Our LNS baseline follows a destroy-and-repair paradigm for the TSPTW inspired by the large neighborhood search framework of \cite{Shaw1998UsingCP}. For each task, we generate an initial tour by sampling 30 random permutations and selecting the best under $\mathrm{score}(\pi)$. The search then performs up to 500 destroy-and-repair iterations, subject to the common evaluation budget. At each iteration, the algorithm destroys the current tour by removing a subset of cities, and repairs the partial tour via greedy insertion. For each removed city, the repair step evaluates all insertion positions and selects the position yielding the lowest $\mathrm{score}(\pi)$ among the tested candidates. The current tour is updated when the repaired tour improves $\mathrm{score}(\pi)$, while the best tour found during the run is tracked and returned.

\subsection{Variable Neighborhood Search (VNS)}

Our VNS baseline follows the general VNS framework for the TSPTW proposed in \cite{DaSilvaUrrutia2010}. Two neighborhood operators are used: \emph{relocate} (remove one city and reinsert it at a different position) and \emph{2-opt} (reverse a subsequence). For each task, we generate an initial tour by sampling 30 random permutations and selecting the best under $\mathrm{score}(\pi)$. The search then performs up to 1000 outer iterations, subject to the common evaluation budget. Each iteration applies a shaking phase consisting of exactly 1 random neighborhood move, followed by a local improvement phase that samples up to 200 neighboring tours and accepts improving moves under $\mathrm{score}(\pi)$. We maintain both a current tour and a global best tour, updating them whenever an improved tour is found, until the evaluation budget is exhausted.

\subsection{LKH-3 Baseline}

We use the original LKH-3 \cite{helsgaun2017extension} implementation as the base solver. Since LKH-3 uses internal termination parameters (e.g., maximum trials) that are not directly comparable to our evaluation budget, we instrument the solver to count fitness evaluations, where one evaluation corresponds to computing $\mathrm{score}(\pi)$ for a candidate tour. The run terminates once the shared evaluation budget is reached. This modification enforces a comparable stopping condition without altering the neighborhood structure or the search logic of LKH-3.

\section{Time Window Generation for Benchmark Instances}
\label{sec:experiments}

While the classical TSP can be efficiently solved via various algorithms for a very large number of cities, TSPTW is much more challenging due to the additional time window constraints. For the experiment, we used the TSPTW benchmark instances from \cite{Langevin1993,DumasTSPTW1995} with problem sizes $n \in \{20,40,60,80,100,150,200\}$.

The instances are commonly studied under travel-cost minimization, where the edge-cost matrix encodes travel times. Each benchmark instance is provided as a multi-task sequence of five related tasks $(T_1,\ldots,T_5)$. For each size $n$, we consider two distinct benchmark instances, denoted $n.10$ and $n.20$. For example, $60.10$ and $60.20$ are different 60-city instances, not variants of the same instance. Across the five tasks, the city set $V$ and travel-time function $d$ are identical. Tasks differ only in their time-window constraints. We write task $T_k$ as
\[
T_k = \Bigl(V, d, \{[a_i^{(k)}, b_i^{(k)}]\}_{i=1}^{n}\Bigr).
\]

Task $T_1$ is the original instance obtained from the source, whereas tasks $T_2, \ldots, T_5$ are variants generated by progressively applying the partial expansion and swap-addition procedures. Algorithms~\ref{Alg:Partial} and~\ref{alg:add-swap} present the pseudocode for the partial expansion and swap-addition procedures, respectively.

\subsection{Partial time-window expansion generator}

The first set of benchmarks is generated by the use of Algorithm~\ref{Alg:Partial}. For each benchmark instance, $T_1$ is the original TSPTW taken directly from the source benchmark, and $T_2, T_3, T_4, T_5$ are generated sequentially by repeatedly relaxing a small part of the current time-window set. More precisely, to construct $T_k$ from $T_{k-1}$, we randomly select between 10\% and 15\% of the non-depot cities and expand only their current time windows, while leaving all remaining windows unchanged. If $[a_i, b_i]$ is the window of city $i$ in $T_{k-1}$, then for each selected city we decrease the lower bound and increase the upper bound by independently sampled random amounts. Let

\begin{algorithm}[t]
\caption{Partial Time Window Expansion Generator}
\label{Alg:Partial}
\KwIn{Feasible tour $\pi$, travel time matrix $D$, original time windows $\{[a_i,b_i]\}$, expansion rate $\rho$}
\KwOut{New time windows $\{[a_i',b_i']\}$}

Randomly select subset $S \subseteq \{2,\dots,n\}$, where $|S| \in [0.1n, 0.15n]$\;

\For{each node $i \in \{2,\dots,n\}$}{
    \eIf{$i \in S$}{
        $\Delta_i \leftarrow b_i - a_i$\;
        $a_i' \leftarrow \max\{0,\, a_i - U(0,\rho)\cdot \Delta_i\}$\;
        $b_i' \leftarrow b_i + U(0,\rho)\cdot \Delta_i$\;
    }{
        $a_i' \leftarrow a_i$, \quad $b_i' \leftarrow b_i$\;
    }
}
Adjust depot closing time to preserve feasibility\;
\end{algorithm}

\[
\ell_i, u_i \sim U\!\left(0, \rho(b_i-a_i)\right)
\]
be independent, where $\rho$ is a parameter that determines the amount of time window expansion. We use $\rho = 0.3$ in all experiments. We then define the new time window as
\[
[a_i', b_i'] = [\max\{0, a_i - \ell_i\}, b_i + u_i].
\]
The depot closing time is then adjusted to preserve feasibility. Repeating this process from $T_1$ to $T_5$ yields a task sequence in which later tasks are progressively more relaxed versions of earlier ones, while the city set and travel-time matrix remain unchanged.

\subsection{Swap-additive time window generator}

For the second set of benchmarks, we generate the instances using Algorithm~\ref{alg:add-swap}. Task $T_1$ is again the original benchmark instance, but $T_2, T_3, T_4,$ and $T_5$ are generated by sequentially rebuilding the time windows around perturbed visitation patterns rather than by monotone relaxation. To construct $T_k$ from $T_{k-1}$, we take a feasible reference tour, copy it, and perform one random swap of two positions (one swap in all reported experiments). We then compute the arrival times induced by the swapped tour and define a new time window for every city centered around its arrival time using Algorithm~\ref{alg:add-swap}. In the general formulation, the window half-width is controlled by the parameter $\delta$. In this study, we instantiate $\delta$ by $\sigma$, the population standard deviation of the arrival times of the swapped tour, i.e., $\delta := \sigma$. Thus, the generated time windows are

\begin{algorithm}[t]
\caption{Swap-Additive Time Window Generator}
\KwIn{Feasible tour $\pi$, travel time matrix $D$, number of swaps $k$, window-size parameter $\delta$}
\KwOut{Set of new time windows $\{[a_i', b_i'] \mid 1 \le i \le n\}$}

Set $\pi' \leftarrow \pi$\;
\For{$j \leftarrow 1$ \KwTo $k$}{
    Select positions $i_1, i_2 \in \{1,\ldots,n\}$ with $i_1 \neq i_2$ uniformly at random\;
    Swap $\pi'[i_1]$ and $\pi'[i_2]$\;
}

Compute arrival times $\{t_i\}_{i=1}^n$ with respect to $\pi'$ using $D$\;

\For{each node $i \in \{1,\ldots,n\}$}{
    $a_i' \leftarrow \max\{0,\, t_i - \delta\}$\;
    $b_i' \leftarrow t_i + \delta$\;
}

\Return{$\{[a_i', b_i'] \mid 1 \le i \le n\}$}\;
\label{alg:add-swap}
\end{algorithm}

\[
[a_i', b_i'] = [\max\{0, t_i - \delta\}, t_i + \delta].
\]
Applying this procedure repeatedly produces a five-task sequence whose tasks still share the same cities and travel-time matrix, but whose temporal constraints are reassigned around slightly different route structures. Unlike the partial expansion setting, this creates a more disruptive sequence in which consecutive tasks remain related, but not through simple progressive relaxation.

We evaluate each local-search algorithm under two protocols.

\subsection{Standard protocol}

Each task $T_k$ is solved independently from scratch. For every run on task $T_k$, the solver starts from its own initialization procedure and searches until the stopping condition is met.

\subsection{Sequential transfer (Iterative protocol)}

Tasks are solved sequentially in the order $T_1 \rightarrow T_2 \rightarrow \cdots \rightarrow T_5$. Within a run, the best tour found for task $T_{k-1}$ is used as the initial tour for task $T_k$. This protocol tests whether reusing tours obtained under earlier related tasks improves feasibility recovery and final performance on subsequent tasks under the same function-evaluation budget.

\section{Experimental Investigations}
\label{sec:results}

\subsection{Experimental Settings}

For algorithm evaluation, we use the original LKH-3 implementation \cite{helsgaun2017extension} written in C, while LNS and VNS are implemented in Python. Each algorithm is independently run 30 times on every task of every benchmark instance. Each run on a given task-instance combination on each instance is terminated after a budget of 100{,}000 function evaluations (FE), that is, 100{,}000 FE are allocated to each instance separately. Experiments were conducted on a Linux-based machine equipped with an AMD Opteron 6348 processor (48 cores, 2.8 GHz) and 120 GB of RAM.

All algorithms are stochastic. Therefore, for each protocol, task, instance, and repetition, we used an independent random seed so that repeated runs explore different search trajectories rather than reproducing the same solution path. This allows us to estimate both average performance and variability over multiple runs.

To assess the statistical significance of the performance differences between the iterative and standard protocols, we applied the non-parametric Mann-Whitney U test at significance level $\alpha = 0.05$ to the penalized scores obtained over 30 independent runs for each task-instance-size combination. Because the number of individual comparisons is large and the paper space is limited, we summarize these outcomes in aggregated figures rather than a full set of per-case tables. In the stacked-bar plots, each bar reports the proportion of transfer-relevant comparisons (Tasks $T_2$--$T_5$) for which the iterative protocol is statistically significantly better, statistically significantly worse, or not statistically significantly different from the standard protocol. The success-rate figures complement this view by showing the mean feasibility rate over the same set of transfer tasks and instances.

For illustration, the main paper includes only two representative results tables for LNS, one for each environment. The remaining tables for all algorithms are provided in Appendix~\ref{sec:supp-tables}. The largest instances ($n = 200$) are omitted from these representative tables. In these tables, \emph{mean} and \emph{std} report the penalized-score mean and standard deviation over 30 runs, \emph{sr} is the feasibility success rate, and \emph{succ.\ $\mu$} and \emph{succ.\ $\sigma$} report the mean and standard deviation over successful runs only. Entries shown as ``--'' indicate that no successful run was obtained for that task-instance combination, and hence these quantities are undefined. The \emph{Stat} column summarizes the Mann--Whitney U outcome: (+) means iterative is significantly better, (-) means iterative is significantly worse, and (*) means no statistically significant difference.

\subsection{Results for the Partial Time Window Expansion Setting}

\begin{figure}[t]
\centering
\begin{tikzpicture}
\begin{groupplot}[
    group style={group size=3 by 1, horizontal sep=0.35cm,
                  ylabels at=edge left,
                  yticklabels at=edge left},
    width=0.30\textwidth,
    height=0.30\textwidth,
    ybar stacked,
    ymin=0, ymax=1,
    ytick={0,0.5,1},
    symbolic x coords={20,40,60,80,100,150,200},
    xtick=data,
    enlarge x limits=0.08,
    ymajorgrids=true,
    grid style={dotted},
    xlabel={Size $n$},
    ylabel={Proportion},
    title style={font=\normalsize},
    tick label style={font=\scriptsize},
    x tick label style={rotate=30, anchor=north east},
    label style={font=\footnotesize},
    legend style={
        font=\footnotesize,
        draw=none,
        at={(0.5,1.18)},
        anchor=south,
        legend columns=3
    }
]

\nextgroupplot[title={LNS}]
\addplot+[draw=black, fill=betterfill] coordinates {
    (20,0) (40,0) (60,0) (80,0.625) (100,0.75) (150,1) (200,1)
};
\addplot+[draw=black, fill=samefill] coordinates {
    (20,1) (40,1) (60,1) (80,0.375) (100,0.25) (150,0) (200,0)
};
\addplot+[draw=black, fill=worsefill] coordinates {
    (20,0) (40,0) (60,0) (80,0) (100,0) (150,0) (200,0)
};

\nextgroupplot[title={VNS}]
\addplot+[draw=black, fill=betterfill] coordinates {
    (20,0) (40,0.75) (60,0.875) (80,1) (100,1) (150,1) (200,1)
};
\addlegendentry{Iterative better}
\addplot+[draw=black, fill=samefill] coordinates {
    (20,1) (40,0.25) (60,0) (80,0) (100,0) (150,0) (200,0)
};
\addlegendentry{No difference}
\addplot+[draw=black, fill=worsefill] coordinates {
    (20,0) (40,0) (60,0.125) (80,0) (100,0) (150,0) (200,0)
};
\addlegendentry{Iterative worse}

\nextgroupplot[title={LKH-3}]
\addplot+[draw=black, fill=betterfill] coordinates {
    (20,0.25) (40,0.125) (60,0.625) (80,0.875) (100,1) (150,1) (200,1)
};
\addplot+[draw=black, fill=samefill] coordinates {
    (20,0.75) (40,0.875) (60,0.25) (80,0.125) (100,0) (150,0) (200,0)
};
\addplot+[draw=black, fill=worsefill] coordinates {
    (20,0) (40,0) (60,0.125) (80,0) (100,0) (150,0) (200,0)
};

\end{groupplot}
\end{tikzpicture}
\caption{Summary of Mann-Whitney U test outcomes for the partial time-window expansion setting across Tasks $T_2$--$T_5$ and both benchmark instances.}
\label{fig:expand-stat-summary}
\end{figure}
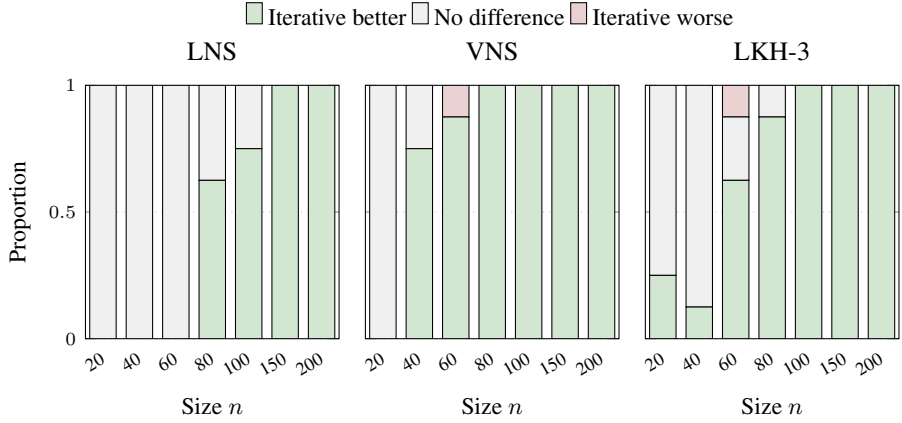

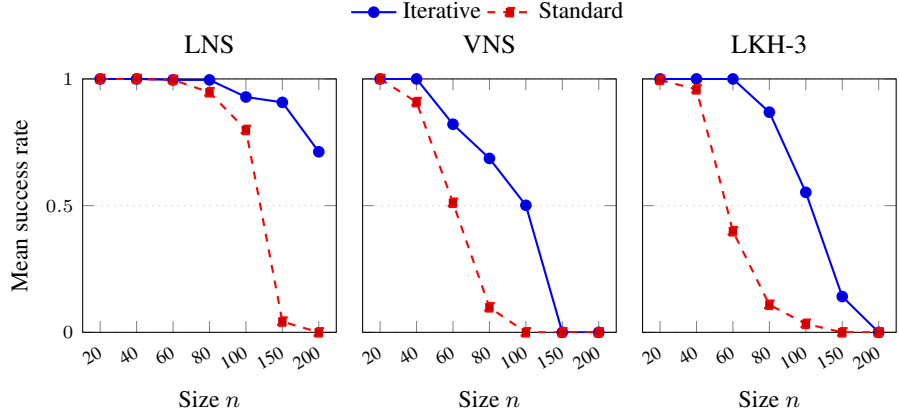
\begin{figure}[t]
\centering
\begin{tikzpicture}
\begin{groupplot}[
    group style={group size=3 by 1, horizontal sep=0.35cm,
                  ylabels at=edge left,
                  yticklabels at=edge left},
    width=0.30\textwidth,
    height=0.30\textwidth,
    ymin=0, ymax=1,
    ytick={0,0.5,1},
    symbolic x coords={20,40,60,80,100,150,200},
    xtick=data,
    enlarge x limits=0.08,
    ymajorgrids=true,
    grid style={dotted},
    xlabel={Size $n$},
    ylabel={Mean success rate},
    title style={font=\normalsize},
    tick label style={font=\scriptsize},
    x tick label style={rotate=30, anchor=north east},
    label style={font=\footnotesize},
    legend style={
        font=\footnotesize,
        draw=none,
        at={(0.5,1.18)},
        anchor=south,
        legend columns=2
    }
]

\nextgroupplot[title={LNS}]
\addplot+[thick, mark=*, mark size=1.8pt] coordinates {
    (20,1) (40,1) (60,0.99625) (80,0.99625) (100,0.92875) (150,0.9075) (200,0.7125)
};
\addplot+[thick, dashed, mark=square*, mark size=1.6pt] coordinates {
    (20,1) (40,1) (60,0.99625) (80,0.9475) (100,0.79875) (150,0.0425) (200,0)
};

\nextgroupplot[title={VNS}]
\addplot+[thick, mark=*, mark size=1.8pt] coordinates {
    (20,1) (40,1) (60,0.82125) (80,0.68625) (100,0.50125) (150,0) (200,0)
};
\addlegendentry{Iterative}
\addplot+[thick, dashed, mark=square*, mark size=1.6pt] coordinates {
    (20,1) (40,0.90875) (60,0.51125) (80,0.09875) (100,0) (150,0) (200,0)
};
\addlegendentry{Standard}

\nextgroupplot[title={LKH-3}]
\addplot+[thick, mark=*, mark size=1.8pt] coordinates {
    (20,1) (40,1) (60,1) (80,0.86875) (100,0.5525) (150,0.14125) (200,0)
};
\addplot+[thick, dashed, mark=square*, mark size=1.6pt] coordinates {
    (20,0.99625) (40,0.95875) (60,0.39875) (80,0.10875) (100,0.03375) (150,0) (200,0)
};

\end{groupplot}
\end{tikzpicture}
\caption{Mean feasibility success rate for the \emph{partial time-window expansion} setting.}
\label{fig:expand-sr-summary}
\end{figure}
\begin{table*}[htbp]
  \tiny
  \centering
  \setlength{\tabcolsep}{2pt}
  \caption{Results under the Partial Time-Window Expansion Setting.}
  \label{tab:expanding}
  \begingroup
  \setlength{\arrayrulewidth}{0pt}
  \resizebox{0.78\textwidth}{!}{%
  \begin{tabular}{ccrrrrr rrrrrc}
      \toprule
      \multirow{2}{*}{\textbf{Instance}} & \multirow{2}{*}{\textbf{Task}} & \multicolumn{5}{c}{\textbf{Iterative LNS}} & \multicolumn{5}{c}{\textbf{Standard LNS}} & \multirow{2}{*}{\textbf{Stat}}\\
      \cmidrule(lr){3-7}\cmidrule(lr){8-12}
       & & \textbf{mean} & \textbf{std} & \textbf{sr} & \textbf{succ. $\mu$} &  \textbf{succ. $\sigma$} & \textbf{mean} & \textbf{std} & \textbf{sr} & \textbf{succ. $\mu$} & \textbf{succ. $\sigma$} & \\ \midrule
      \multirow{5}{*}{20.10}
   & $T_1$ & 661.6 & 0 & 1 & 661.6 & 0 & 661.6 & 0 & 1 & 661.6 & 0 & (*) \\ & $T_2$ & 629.2 & 0 & 1 & 629.2 & 0 & 629.2 & 0 & 1 & 629.2 & 0 & (*) \\ & $T_3$ & 629.2 & 0 & 1 & 629.2 & 0 & 629.2 & 0 & 1 & 629.2 & 0 & (*) \\ & $T_4$ & 629.2 & 0 & 1 & 629.2 & 0 & 629.2 & 0 & 1 & 629.2 & 0 & (*) \\ & $T_5$ & 629.2 & 0 & 1 & 629.2 & 0 & 629.2 & 0 & 1 & 629.2 & 0 & (*) \\    \hline
      \multirow{5}{*}{20.20}
   & $T_1$ & 684.2 & 0 & 1 & 684.2 & 0 & 684.2 & 0 & 1 & 684.2 & 0 & (*) \\ & $T_2$ & 684.2 & 0 & 1 & 684.2 & 0 & 684.2 & 0 & 1 & 684.2 & 0 & (*) \\ & $T_3$ & 684.2 & 0 & 1 & 684.2 & 0 & 684.2 & 0 & 1 & 684.2 & 0 & (*) \\ & $T_4$ & 684.2 & 0 & 1 & 684.2 & 0 & 684.2 & 0 & 1 & 684.2 & 0 & (*) \\ & $T_5$ & 581.8 & 0 & 1 & 581.8 & 0 & 581.8 & 0 & 1 & 581.8 & 0 & (*) \\    \hline
      \multirow{5}{*}{40.10}
   & $T_1$ & 1100.6 & 0 & 1 & 1100.6 & 0 & 1100.6 & 0 & 1 & 1100.6 & 0 & (*) \\ & $T_2$ & 1084.3 & 0 & 1 & 1084.3 & 0 & 1084.3 & 0 & 1 & 1084.3 & 0 & (*) \\ & $T_3$ & 1083.6 & 0 & 1 & 1083.6 & 0 & 1083.6 & 0 & 1 & 1083.6 & 0 & (*) \\ & $T_4$ & 1076.5 & 0 & 1 & 1076.5 & 0 & 1076.5 & 0 & 1 & 1076.5 & 0 & (*) \\ & $T_5$ & 1043.8 & 0 & 1 & 1043.8 & 0 & 1043.8 & 0 & 1 & 1043.8 & 0 & (*) \\    \hline
      \multirow{5}{*}{40.20}
   & $T_1$ & 1010.4 & 0 & 1 & 1010.4 & 0 & 1010.4 & 0 & 1 & 1010.4 & 0 & (*) \\ & $T_2$ & 983.3 & 0 & 1 & 983.3 & 0 & 983.3 & 0 & 1 & 983.3 & 0 & (*) \\ & $T_3$ & 983.2 & 0 & 1 & 983.2 & 0 & 983.2 & 0 & 1 & 983.2 & 0 & (*) \\ & $T_4$ & 956.5 & 0 & 1 & 956.5 & 0 & 956.5 & 0 & 1 & 956.5 & 0 & (*) \\ & $T_5$ & 956.5 & 0 & 1 & 956.5 & 0 & 956.5 & 0 & 1 & 956.5 & 0 & (*) \\    \hline
      \multirow{5}{*}{60.10}
   & $T_1$ & 3.68e5 & 1.55e6 & 0.93 & 591.82 & 1.22 & 3.68e5 & 1.55e6 & 0.93 & 591.82 & 1.22 & (*) \\ & $T_2$ & 2.29e5 & 1.25e6 & 0.97 & 504.41 & 0.5 & 504.57 & 0.94 & 1 & 504.57 & 0.94 & (*) \\ & $T_3$ & 503.4 & 0.81 & 1 & 503.4 & 0.81 & 504.3 & 3.21 & 1 & 504.3 & 3.21 & (*) \\ & $T_4$ & 471.4 & 0.81 & 1 & 471.4 & 0.81 & 473.03 & 8.92 & 1 & 473.03 & 8.92 & (*) \\ & $T_5$ & 469.13 & 1.38 & 1 & 469.13 & 1.38 & 4.88e4 & 2.65e5 & 0.97 & 469.79 & 10.05 & (*) \\    \hline
      \multirow{5}{*}{60.20}
   & $T_1$ & 1353.5 & 0 & 1 & 1353.5 & 0 & 1353.5 & 0 & 1 & 1353.5 & 0 & (*) \\ & $T_2$ & 1353.5 & 0 & 1 & 1353.5 & 0 & 1353.5 & 0 & 1 & 1353.5 & 0 & (*) \\ & $T_3$ & 1349.1 & 0 & 1 & 1349.1 & 0 & 1349.1 & 0 & 1 & 1349.1 & 0 & (*) \\ & $T_4$ & 1263 & 0 & 1 & 1263 & 0 & 1263 & 0 & 1 & 1263 & 0 & (*) \\ & $T_5$ & 1205.6 & 0 & 1 & 1205.6 & 0 & 1205.6 & 0 & 1 & 1205.6 & 0 & (*) \\    \hline
      \multirow{5}{*}{80.10}
   & $T_1$ & 2.26e5 & 1.24e6 & 0.97 & 619.07 & 2.02 & 2.26e5 & 1.24e6 & 0.97 & 619.07 & 2.02 & (*) \\ & $T_2$ & 2.26e5 & 1.24e6 & 0.97 & 614.76 & 1.12 & 2.26e5 & 1.24e6 & 0.97 & 614.55 & 1.59 & (*) \\ & $T_3$ & 597.67 & 2.87 & 1 & 597.67 & 2.87 & 599.1 & 3.52 & 1 & 599.1 & 3.52 & (+) \\ & $T_4$ & 575.73 & 1.96 & 1 & 575.73 & 1.96 & 577.97 & 3.51 & 1 & 577.97 & 3.51 & (+) \\ & $T_5$ & 574.03 & 2.82 & 1 & 574.03 & 2.82 & 1.11e4 & 5.75e4 & 0.97 & 581.1 & 4.26 & (+) \\    \hline
      \multirow{5}{*}{80.20}
   & $T_1$ & 1.72e4 & 9.04e4 & 0.97 & 677.03 & 2.96 & 1.72e4 & 9.04e4 & 0.97 & 677.03 & 2.96 & (*) \\ & $T_2$ & 669.17 & 3.9 & 1 & 669.17 & 3.9 & 1.05e5 & 3.23e5 & 0.9 & 671.41 & 5.94 & (+) \\ & $T_3$ & 653.73 & 1.34 & 1 & 653.73 & 1.34 & 8.32e4 & 2.56e5 & 0.9 & 653.78 & 2.86 & (*) \\ & $T_4$ & 625.83 & 5.25 & 1 & 625.83 & 5.25 & 1.82e5 & 5.52e5 & 0.87 & 619.12 & 7.78 & (+) \\ & $T_5$ & 604.63 & 9.5 & 1 & 604.63 & 9.5 & 2.26e5 & 1.24e6 & 0.97 & 605.76 & 16.44 & (*) \\    \hline
      \multirow{5}{*}{100.10}
   & $T_1$ & 8.43e5 & 1.33e6 & 0.63 & 771.37 & 1.38 & 8.43e5 & 1.33e6 & 0.63 & 771.37 & 1.38 & (*) \\ & $T_2$ & 8.43e5 & 1.33e6 & 0.63 & 760.32 & 1.42 & 9.44e6 & 3.34e7 & 0.63 & 760.42 & 1.61 & (*) \\ & $T_3$ & 5.07e4 & 1.02e5 & 0.8 & 701.21 & 2.62 & 1.26e7 & 3.30e7 & 0.6 & 709.39 & 7.01 & (+) \\ & $T_4$ & 694.57 & 8.02 & 1 & 694.57 & 8.02 & 1.21e7 & 3.35e7 & 0.73 & 692.41 & 8.66 & (*) \\ & $T_5$ & 683.73 & 8.47 & 1 & 683.73 & 8.47 & 1.05e7 & 2.67e7 & 0.6 & 684.28 & 15.64 & (+) \\    \hline
      \multirow{5}{*}{100.20}
   & $T_1$ & 665.53 & 9.04 & 1 & 665.53 & 9.04 & 665.53 & 9.04 & 1 & 665.53 & 9.04 & (*) \\ & $T_2$ & 651.7 & 9.66 & 1 & 651.7 & 9.66 & 5.07e4 & 2.74e5 & 0.97 & 661.48 & 12.54 & (+) \\ & $T_3$ & 628.17 & 6.43 & 1 & 628.17 & 6.43 & 9.23e4 & 3.50e5 & 0.93 & 635.75 & 10.3 & (+) \\ & $T_4$ & 626.03 & 4.12 & 1 & 626.03 & 4.12 & 634.97 & 9.6 & 1 & 634.97 & 9.6 & (+) \\ & $T_5$ & 618.73 & 2.97 & 1 & 618.73 & 2.97 & 1.30e6 & 5.40e6 & 0.93 & 626.82 & 11.15 & (+) \\    \hline
      \multirow{5}{*}{150.10}
   & $T_1$ & 4.91e8 & 5.14e8 & 0.07 & 969 & 24.04 & 4.91e8 & 5.14e8 & 0.07 & 969 & 24.04 & (*) \\ & $T_2$ & 2.69e7 & 1.02e8 & 0.93 & 938.5 & 5.06 & 5.03e8 & 3.75e8 & 0.07 & 945.5 & 9.19 & (+) \\ & $T_3$ & 906.2 & 3.55 & 1 & 906.2 & 3.55 & 5.21e8 & 4.32e8 & 0 & -- & -- & (+) \\ & $T_4$ & 905.53 & 2.99 & 1 & 905.53 & 2.99 & 5.29e8 & 4.59e8 & 0.03 & 941 & 0 & (+) \\ & $T_5$ & 902 & 3.46 & 1 & 902 & 3.46 & 5.31e8 & 3.45e8 & 0 & -- & -- & (+) \\    \hline
      \multirow{5}{*}{150.20}
   & $T_1$ & 4.16e8 & 4.07e8 & 0.1 & 748 & 3 & 4.16e8 & 4.07e8 & 0.1 & 748 & 3 & (*) \\ & $T_2$ & 1.33e7 & 6.56e7 & 0.73 & 724.32 & 5.2 & 4.30e8 & 4.22e8 & 0.07 & 753 & 1.41 & (+) \\ & $T_3$ & 6.97e5 & 2.87e6 & 0.83 & 698.84 & 4.58 & 3.78e8 & 2.66e8 & 0.07 & 732.5 & 19.09 & (+) \\ & $T_4$ & 5.81e5 & 2.65e6 & 0.87 & 691.5 & 4.22 & 4.20e8 & 3.89e8 & 0.07 & 731.5 & 16.26 & (+) \\ & $T_5$ & 5.62e5 & 2.65e6 & 0.9 & 624.59 & 4.01 & 4.95e8 & 4.25e8 & 0.03 & 734 & 0 & (+) \\    \hline
  \bottomrule
  \end{tabular}%
  }
  \endgroup
\end{table*}

We first evaluate the two protocols on the partial time-window expansion setting, where each successive task relaxes the time-window constraints while preserving the overall task order. This is the more transfer-aligned environment, since a tour obtained on an earlier task remains structurally relevant for later tasks.

Table~\ref{tab:expanding} provides representative task-level evidence for LNS under the partial time-window expansion setting and mirrors the size-dependent pattern observed in Figures~\ref{fig:expand-stat-summary} and~\ref{fig:expand-sr-summary}. On the smaller instances, particularly $60.10$ and $60.20$, most comparisons over the transfer tasks $T_2$--$T_5$ are not statistically significant, indicating that when both protocols already solve the tasks reliably, there is limited room for transfer to improve performance. From $n = 80$ onward, however, the iterative protocol becomes increasingly favorable. On instance $80.10$, it is already significantly better on $T_3$--$T_5$, and on larger instances such as $150.10$, $150.20$, $200.10$, and $200.20$, the advantage becomes pronounced, with markedly higher feasibility rates and substantially lower penalized scores on most transfer tasks. These task-level results reinforce the aggregate evidence that when the task sequence is generated by progressive relaxation, initializing task $T_k$ from the best solution obtained for $T_{k-1}$ improves both feasibility recovery and final solution quality, especially as the instances become more difficult.

Figure~\ref{fig:expand-stat-summary} summarizes the statistical outcomes over the transfer tasks $T_2$--$T_5$. Overall, a clear size-dependent pattern can be observed. For small instances, especially $n=20$, the iterative and standard protocols are often not statistically significantly different. However, as the instance size increases, the iterative protocol becomes increasingly favorable. This trend is particularly strongest for VNS and LKH-3, where from medium sizes onward, most comparisons indicate that the iterative protocol is statistically significantly better than the standard protocol. For LNS, the same pattern emerges more gradually, with the advantage becoming pronounced from $n=80$ onward.

For each algorithm and problem size, Figure~\ref{fig:expand-sr-summary} reports the average proportion of runs that reached a feasible tour across Tasks $T_2$--$T_5$ and both benchmark instances, separately for the iterative and standard protocols. The success-rate trends support the same conclusion. For LNS, both protocols remain highly successful on small and medium instances, but the iterative protocol maintains substantially higher feasibility rates once the instances become harder, particularly at $n=150$ and $n=200$. The effect is even more noticeable for VNS and LKH-3. In both cases, success rates under the standard protocol deteriorate much earlier as $n$ grows, whereas the iterative protocol preserves a markedly higher probability of finding feasible tours over a wider range of instance sizes.

These results indicate that the iterative protocol is especially beneficial when the task sequence is generated by progressive relaxation. In this setting, the solution found for task $T_{k-1}$ provides a useful starting point for task $T_k$, improving both the ability to recover feasible tours and the final penalized score. Another notable observation is that the benefit becomes more visible as the search space becomes harder to explore from scratch. On small instances, both protocols often reach similarly good solutions, leaving little room for improvement through transfer. On medium and large instances, however, the iterative protocol more consistently exploits cross-task similarity and delivers stronger performance.

Overall, the partial expansion results provide strong evidence that the iterative protocol is preferable to solving each task independently from scratch when the sequence of tasks is constructed through gradual relaxation. This motivates the next experiment, where we investigate whether the same advantage persists under a more disruptive setting in which the time windows are modified by swapping.

\subsection{Results for the Swap-Additive Time Window Setting}

To further test the robustness of sequential transfer, we next consider the more demanding swap-additive time-window setting. In this environment, tasks are not produced through gradual relaxation. Instead, each new task is generated by perturbing a feasible reference tour with a swap and rebuilding the time windows around the arrival times induced by that perturbed tour. This weakens the structural continuity between consecutive tasks and makes iterative problem solving less straightforward.

\begin{table*}[htbp]
  \tiny
  \centering
  \setlength{\tabcolsep}{2pt}
  \caption{Results under the Swap-Additive Time-Window Setting.}
  \label{tab:swap}
  \begingroup
  \setlength{\arrayrulewidth}{0pt}
  \resizebox{0.80\textwidth}{!}{%
  \begin{tabular}{ccrrrrr rrrrrc}
      \toprule
      \multirow{2}{*}{\textbf{Instance}} & \multirow{2}{*}{\textbf{Task}} & \multicolumn{5}{c}{\textbf{Iterative LNS}} & \multicolumn{5}{c}{\textbf{Standard LNS}} & \multirow{2}{*}{\textbf{Stat}}\\
      \cmidrule(lr){3-7}\cmidrule(lr){8-12}
       & & \textbf{mean} & \textbf{std} & \textbf{sr} & \textbf{succ. $\mu$} &  \textbf{succ. $\sigma$} & \textbf{mean} & \textbf{std} & \textbf{sr} & \textbf{succ. $\mu$} & \textbf{succ. $\sigma$} & \\ \midrule
      \multirow{5}{*}{20.10}
   & $T_1$ & 713.20 & 0 & 1 & 713.20 & 0 & 713.20 & 0 & 1 & 713.20 & 0 & (*) \\ & $T_2$ & 706.30 & 0 & 1 & 706.30 & 0 & 706.30 & 0 & 1 & 706.30 & 0 & (*) \\ & $T_3$ & 798.40 & 0 & 1 & 798.40 & 0 & 798.40 & 0 & 1 & 798.40 & 0 & (*) \\ & $T_4$ & 919.50 & 0 & 1 & 919.50 & 0 & 919.50 & 0 & 1 & 919.50 & 0 & (*) \\ & $T_5$ & 934.20 & 0 & 1 & 934.20 & 0 & 934.20 & 0 & 1 & 934.20 & 0 & (*) \\    \hline
      \multirow{5}{*}{20.20}
   & $T_1$ & 781 & 0 & 1 & 781 & 0 & 781 & 0 & 1 & 781 & 0 & (*) \\ & $T_2$ & 831 & 0 & 1 & 831 & 0 & 831 & 0 & 1 & 831 & 0 & (*) \\ & $T_3$ & 842.40 & 0 & 1 & 842.40 & 0 & 842.40 & 0 & 1 & 842.40 & 0 & (*) \\ & $T_4$ & 856.20 & 0 & 1 & 856.20 & 0 & 856.20 & 0 & 1 & 856.20 & 0 & (*) \\ & $T_5$ & 894.90 & 0 & 1 & 894.90 & 0 & 894.90 & 0 & 1 & 894.90 & 0 & (*) \\    \hline
      \multirow{5}{*}{40.10}
   & $T_1$ & 1429 & 0 & 1 & 1429 & 0 & 1429 & 0 & 1 & 1429 & 0 & (*) \\ & $T_2$ & 1521.80 & 0 & 1 & 1521.80 & 0 & 1523.54 & 2.70 & 1 & 1523.54 & 2.70 & (+) \\ & $T_3$ & 1492.18 & 2.89 & 1 & 1492.18 & 2.89 & 1492.47 & 2.90 & 1 & 1492.47 & 2.90 & (*) \\ & $T_4$ & 1596.32 & 3.30 & 1 & 1596.32 & 3.30 & 1596.03 & 2.96 & 1 & 1596.03 & 2.96 & (*) \\ & $T_5$ & 1739.01 & 5.34 & 1 & 1739.01 & 5.34 & 1740.04 & 4.65 & 1 & 1740.04 & 4.65 & (*) \\    \hline
      \multirow{5}{*}{40.20}
   & $T_1$ & 1320.92 & 4.98 & 1 & 1320.92 & 4.98 & 1320.92 & 4.98 & 1 & 1320.92 & 4.98 & (*) \\ & $T_2$ & 1360.50 & 0 & 1 & 1360.50 & 0 & 1360.50 & 0 & 1 & 1360.50 & 0 & (*) \\ & $T_3$ & 1419.20 & 0 & 1 & 1419.20 & 0 & 1419.20 & 0 & 1 & 1419.20 & 0 & (*) \\ & $T_4$ & 1438.90 & 0 & 1 & 1438.90 & 0 & 1438.90 & 0 & 1 & 1438.90 & 0 & (*) \\ & $T_5$ & 1511.80 & 0 & 1 & 1511.80 & 0 & 1511.80 & 0 & 1 & 1511.80 & 0 & (*) \\    \hline
      \multirow{5}{*}{60.10}
   & $T_1$ & 1596.16 & 3.05 & 1 & 1596.16 & 3.05 & 1596.16 & 3.05 & 1 & 1596.16 & 3.05 & (*) \\ & $T_2$ & 1632.31 & 1.98 & 1 & 1632.31 & 1.98 & 1632.41 & 1.95 & 1 & 1632.41 & 1.95 & (*) \\ & $T_3$ & 1626.85 & 0.14 & 1 & 1626.85 & 0.14 & 1626.84 & 0.12 & 1 & 1626.84 & 0.12 & (*) \\ & $T_4$ & 1711.87 & 0.15 & 1 & 1711.87 & 0.15 & 1712.33 & 1.60 & 1 & 1712.33 & 1.60 & (*) \\ & $T_5$ & 1801.59 & 1.57 & 1 & 1801.59 & 1.57 & 1801.80 & 1.33 & 1 & 1801.80 & 1.33 & (*) \\    \hline
      \multirow{5}{*}{60.20}
   & $T_1$ & 3897.93 & 1.76e4 & 0.97 & 676.17 & 6.47 & 3897.93 & 1.76e4 & 0.97 & 676.17 & 6.47 & (*) \\ & $T_2$ & 699.43 & 4.41 & 1 & 699.43 & 4.41 & 701.63 & 7.14 & 1 & 701.63 & 7.14 & (*) \\ & $T_3$ & 732.70 & 4.05 & 1 & 732.70 & 4.05 & 735.33 & 6.40 & 1 & 735.33 & 6.40 & (*) \\ & $T_4$ & 731.97 & 3.77 & 1 & 731.97 & 3.77 & 735.33 & 6.40 & 1 & 735.33 & 6.40 & (+) \\ & $T_5$ & 731.57 & 3.45 & 1 & 731.57 & 3.45 & 734.00 & 6.26 & 1 & 734.00 & 6.26 & (*) \\    \hline
      \multirow{5}{*}{80.10}
   & $T_1$ & 3.76e4 & 2.01e5 & 0.97 & 865.07 & 7.44 & 3.76e4 & 2.01e5 & 0.97 & 865.07 & 7.44 & (*) \\ & $T_2$ & 862.47 & 5.65 & 1 & 862.47 & 5.65 & 863.90 & 6.64 & 1 & 863.90 & 6.64 & (*) \\ & $T_3$ & 868.10 & 5.92 & 1 & 868.10 & 5.92 & 868.87 & 4.71 & 1 & 868.87 & 4.71 & (*) \\ & $T_4$ & 851.00 & 4.65 & 1 & 851.00 & 4.65 & 852.73 & 7.91 & 1 & 852.73 & 7.91 & (*) \\ & $T_5$ & 853.03 & 3.58 & 1 & 853.03 & 3.58 & 859.00 & 5.34 & 1 & 859.00 & 5.34 & (+) \\    \hline
      \multirow{5}{*}{80.20}
   & $T_1$ & 971.17 & 5.25 & 1 & 971.17 & 5.25 & 971.17 & 5.25 & 1 & 971.17 & 5.25 & (*) \\ & $T_2$ & 977.27 & 12.68 & 1 & 977.27 & 12.68 & 982.10 & 13.45 & 1 & 982.10 & 13.45 & (*) \\ & $T_3$ & 990.00 & 16.33 & 1 & 990.00 & 16.33 & 991.23 & 14.00 & 1 & 991.23 & 14.00 & (*) \\ & $T_4$ & 2.3e4 & 1.21e5 & 0.97 & 1012.52 & 9.59 & 1015.67 & 17.25 & 1 & 1015.67 & 17.25 & (*) \\ & $T_5$ & 1101.27 & 13.97 & 1 & 1101.27 & 13.97 & 1100.90 & 13.93 & 1 & 1100.90 & 13.93 & (*) \\    \hline
       \multirow{5}{*}{100.10}
    & $T_1$ & 6.76e5 & 1.31e6 & 0.77 & 1273.61 & 17.89 & 6.76e5 & 1.31e6 & 0.77 & 1273.61 & 17.89 & (*) \\ & $T_2$ & 1.99e6 & 4.95e6 & 0.80 & 1302.79 & 21.86 & 2.32e6 & 4.04e6 & 0.57 & 1302.94 & 21.48 & (*) \\ & $T_3$ & 1.26e5 & 4.86e5 & 0.93 & 1297.00 & 13.47 & 1.8e6 & 3.04e6 & 0.50 & 1306.20 & 15.41 & (+) \\ & $T_4$ & 3.6e5 & 1.96e6 & 0.97 & 1303.62 & 16.58 & 5.51e5 & 9.43e5 & 0.73 & 1313.41 & 17.46 & (+) \\ & $T_5$ & 4.43e5 & 1.97e6 & 0.90 & 1252.15 & 22.28 & 2.01e5 & 5.55e5 & 0.83 & 1255.24 & 17.38 & (*) \\    \hline
       \multirow{5}{*}{100.20}
    & $T_1$ & 3.51e5 & 1.07e6 & 0.90 & 1064.00 & 18.52 & 3.51e5 & 1.07e6 & 0.90 & 1064.00 & 18.52 & (*) \\ & $T_2$ & 1.18e5 & 6.39e5 & 0.97 & 1067.10 & 14.26 & 3.51e5 & 1.07e6 & 0.90 & 1071.11 & 16.43 & (*) \\ & $T_3$ & 1104.93 & 22.64 & 1 & 1104.93 & 22.64 & 1110.57 & 30.76 & 1 & 1110.57 & 30.76 & (*) \\ & $T_4$ & 8.45e4 & 4.56e5 & 0.97 & 1186.00 & 18.49 & 1.68e5 & 6.34e5 & 0.93 & 1203.75 & 33.16 & (*) \\ & $T_5$ & 1175.03 & 20.00 & 1 & 1175.03 & 20.00 & 1.35e5 & 5.07e5 & 0.93 & 1194.04 & 30.10 & (+) \\    \hline
       \multirow{5}{*}{150.10}
    & $T_1$ & 5.47e8 & 5.9e8 & 0.17 & 1372.00 & 28.71 & 5.47e8 & 5.9e8 & 0.17 & 1372.00 & 28.71 & (*) \\ & $T_2$ & 2.31e6 & 8.55e6 & 0.80 & 1364.83 & 21.99 & 5.79e8 & 6.94e8 & 0.17 & 1399.00 & 20.41 & (+) \\ & $T_3$ & 1.51e8 & 2.54e8 & 0.40 & 1448.67 & 44.63 & 6.84e8 & 7.76e8 & 0.20 & 1451.67 & 22.80 & (+) \\ & $T_4$ & 1.45e8 & 1.5e8 & 0.17 & 1442.00 & 53.19 & 6.89e8 & 8.56e8 & 0.23 & 1428.86 & 35.13 & (*) \\ & $T_5$ & 8.98e7 & 1.53e8 & 0.63 & 1452.58 & 39.63 & 8.46e8 & 6.38e8 & 0.13 & 1452.00 & 35.19 & (+) \\    \hline
       \multirow{5}{*}{150.20}
    & $T_1$ & 5.55e8 & 6.5e8 & 0.17 & 1344.60 & 38.58 & 5.55e8 & 6.5e8 & 0.17 & 1344.60 & 38.58 & (*) \\ & $T_2$ & 1.43e6 & 7.1e6 & 0.90 & 1280.93 & 33.35 & 6.03e8 & 6.25e8 & 0.17 & 1301.20 & 28.04 & (+) \\ & $T_3$ & 1.16e6 & 6.36e6 & 0.97 & 1277.48 & 24.96 & 6.07e8 & 6.29e8 & 0.17 & 1321.40 & 8.82 & (+) \\ & $T_4$ & 1.12e6 & 6.15e6 & 0.97 & 1222.55 & 27.11 & 6.5e8 & 6.83e8 & 0.17 & 1292.60 & 22.27 & (+) \\ & $T_5$ & 2.51e8 & 3.43e8 & 0.23 & 1407.71 & 74.09 & 7.07e8 & 7.54e8 & 0.13 & 1363.75 & 31.34 & (+) \\    \hline
  \bottomrule
  \end{tabular}%
  }
  \endgroup
\end{table*}

Table~\ref{tab:swap} shows the corresponding task-level picture for LNS under the swap-additive time-window setting, where the continuity between consecutive tasks is weaker, and the transfer mechanism is therefore less directly aligned with the task-generation process. In this setting, the benefit of the iterative protocol is more mixed than in the partial-expansion environment, and there is no feasible solution for instances with size larger than 100. Across many comparisons over $T_2$--$T_5$, especially on the smaller instances, the two protocols are not statistically significantly different. At the same time, several comparisons favor the iterative protocol through lower penalized scores. Overall, Table~\ref{tab:swap} suggests that under swap-additive changes, the iterative protocol is competitive, but its advantage is less uniform than under progressive relaxation.

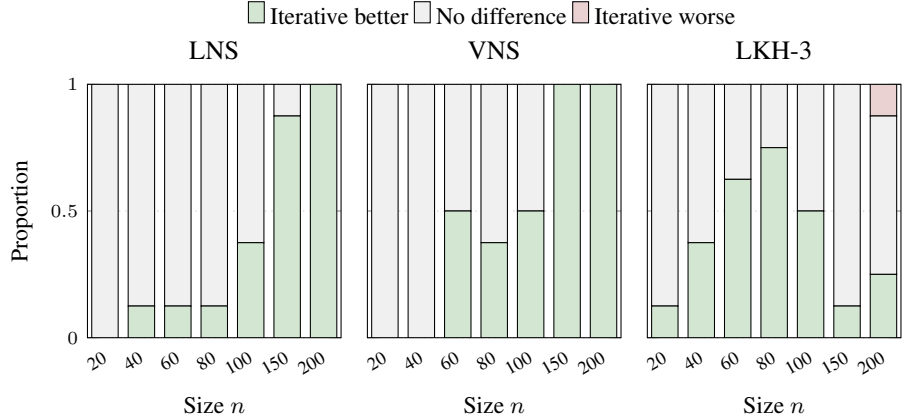
\begin{figure}[t]
\centering
\begin{tikzpicture}
\begin{groupplot}[
    group style={group size=3 by 1, horizontal sep=0.35cm,
                  ylabels at=edge left,
                  yticklabels at=edge left},
    width=0.30\textwidth,
    height=0.30\textwidth,
    ybar stacked,
    ymin=0, ymax=1,
    ytick={0,0.5,1},
    symbolic x coords={20,40,60,80,100,150,200},
    xtick=data,
    enlarge x limits=0.08,
    ymajorgrids=true,
    grid style={dotted},
    xlabel={Size $n$},
    ylabel={Proportion},
    title style={font=\normalsize},
    tick label style={font=\scriptsize},
    x tick label style={rotate=30, anchor=north east},
    label style={font=\footnotesize},
    legend style={
        font=\footnotesize,
        draw=none,
        at={(0.5,1.18)},
        anchor=south,
        legend columns=3
    }
]

\nextgroupplot[title={LNS}]
\addplot+[draw=black, fill=betterfill] coordinates {
    (20,0) (40,0.125) (60,0.125) (80,0.125) (100,0.375) (150,0.875) (200,1)
};
\addplot+[draw=black, fill=samefill] coordinates {
    (20,1) (40,0.875) (60,0.875) (80,0.875) (100,0.625) (150,0.125) (200,0)
};
\addplot+[draw=black, fill=worsefill] coordinates {
    (20,0) (40,0) (60,0) (80,0) (100,0) (150,0) (200,0)
};

\nextgroupplot[title={VNS}]
\addplot+[draw=black, fill=betterfill] coordinates {
    (20,0) (40,0) (60,0.5) (80,0.375) (100,0.5) (150,1) (200,1)
};
\addlegendentry{Iterative better}
\addplot+[draw=black, fill=samefill] coordinates {
    (20,1) (40,1) (60,0.5) (80,0.625) (100,0.5) (150,0) (200,0)
};
\addlegendentry{No difference}
\addplot+[draw=black, fill=worsefill] coordinates {
    (20,0) (40,0) (60,0) (80,0) (100,0) (150,0) (200,0)
};
\addlegendentry{Iterative worse}

\nextgroupplot[title={LKH-3}]
\addplot+[draw=black, fill=betterfill] coordinates {
    (20,0.125) (40,0.375) (60,0.625) (80,0.75) (100,0.5) (150,0.125) (200,0.25)
};
\addplot+[draw=black, fill=samefill] coordinates {
    (20,0.875) (40,0.625) (60,0.375) (80,0.25) (100,0.5) (150,0.875) (200,0.625)
};
\addplot+[draw=black, fill=worsefill] coordinates {
    (20,0) (40,0) (60,0) (80,0) (100,0) (150,0) (200,0.125)
};

\end{groupplot}
\end{tikzpicture}
\caption{Summary of Mann-Whitney U test outcomes for the swap-additive time-window setting across Tasks $T_2$--$T_5$ and both benchmark instances.}
\label{fig:swap-stat-summary}
\end{figure}

Figure~\ref{fig:swap-stat-summary} summarizes the proportion of Mann-Whitney U test outcomes computed over Tasks $T_2$--$T_5$ and both benchmark instances in which the iterative protocol performs significantly better, significantly worse, or shows no significant difference relative to the standard protocol. As Figure~\ref{fig:swap-stat-summary} indicates, the effect of the iterative protocol is noticeably less consistent than in the partial expansion setting. On smaller instances, most comparisons show no significant statistical difference between the two protocols, suggesting that both approaches handle these easier cases similarly. As the problem size increases, however, positive outcomes for the iterative protocol become more frequent for LNS and VNS, with the clearest gains appearing on harder instances. LKH-3 follows a less stable pattern, although the iterative protocol remains broadly competitive across most settings.

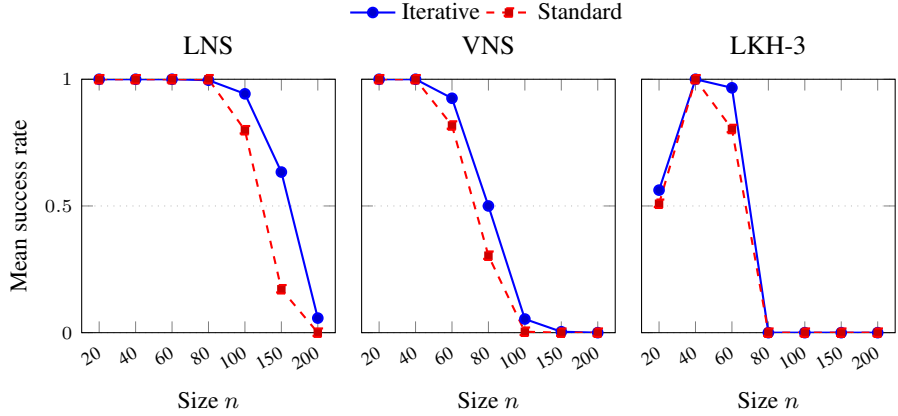
\begin{figure}[t]
    \centering
    \begin{tikzpicture}
    \begin{groupplot}[
        group style={group size=3 by 1, horizontal sep=0.35cm,
                      ylabels at=edge left,
                      yticklabels at=edge left},
        width=0.30\textwidth,
        height=0.30\textwidth,
        ymin=0, ymax=1,
        ytick={0,0.5,1},
        symbolic x coords={20,40,60,80,100,150,200},
        xtick=data,
        enlarge x limits=0.08,
        ymajorgrids=true,
        grid style={dotted},
        xlabel={Size $n$},
        ylabel={Mean success rate},
        title style={font=\normalsize},
        tick label style={font=\scriptsize},
        x tick label style={rotate=30, anchor=north east},
        label style={font=\footnotesize},
        legend style={
            font=\footnotesize,
            draw=none,
            at={(0.5,1.18)},
            anchor=south,
            legend columns=2
        }
    ]
    
    \nextgroupplot[title={LNS}]
    \addplot+[thick, mark=*, mark size=1.8pt] coordinates {
        (20,1) (40,1) (60,1) (80,0.99625) (100,0.9425) (150,0.63375) (200,0.0575)
    };
    \addplot+[thick, dashed, mark=square*, mark size=1.6pt] coordinates {
        (20,1) (40,1) (60,1) (80,1) (100,0.79875) (150,0.17125) (200,0)
    };
    
    \nextgroupplot[title={VNS}]
    \addplot+[thick, mark=*, mark size=1.8pt] coordinates {
        (20,1) (40,1) (60,0.925) (80,0.5) (100,0.05375) (150,0.00375) (200,0)
    };
    \addlegendentry{Iterative}
    \addplot+[thick, dashed, mark=square*, mark size=1.6pt] coordinates {
        (20,1) (40,1) (60,0.8175) (80,0.30375) (100,0.00375) (150,0) (200,0)
    };
    \addlegendentry{Standard}
    
    \nextgroupplot[title={LKH-3}]
    \addplot+[thick, mark=*, mark size=1.8pt] coordinates {
        (20,0.5625) (40,1) (60,0.96625) (80,0) (100,0) (150,0) (200,0)
    };
    \addplot+[thick, dashed, mark=square*, mark size=1.6pt] coordinates {
        (20,0.50875) (40,1) (60,0.80375) (80,0) (100,0) (150,0) (200,0)
    };
    
    \end{groupplot}
    \end{tikzpicture}
    \caption{Mean feasibility success rate for the \emph{swap-additive time-window} setting.}
    \label{fig:swap-sr-summary}
    \end{figure}

Figure~\ref{fig:swap-sr-summary} presents, for each algorithm and problem size, the average proportion of runs that achieved a feasible tour under the swap-additive time-window setting. These results are aggregated across Tasks $T_2$--$T_5$ and both benchmark instances, and are reported separately for the iterative and standard evaluation protocols. This is consistent with the statistical-outcome summary in Figure~\ref{fig:swap-stat-summary}. Under swapping, feasibility becomes more difficult to sustain as the instances become larger, and both protocols begin to degrade earlier than they do in the partial expansion environment. Nevertheless, the iterative protocol generally achieves equal or higher success rates than the standard protocol, particularly for LNS and VNS on medium-to-large instances. This indicates that, despite the disrupted task similarity, initializing from the previous task can still guide the search in a useful direction.

In summary, these findings suggest that the gains from iteration are weaker and less uniform in the swapping setting, which is consistent with the reduced relationship between consecutive tasks. Many comparisons are statistically indistinguishable, while some harder-instance settings favor the iterative protocol. Overall, the iterative protocol is best characterized here as broadly competitive, with occasional benefits, rather than uniformly superior.

\section{Conclusions}
\label{sec:conclusion}

We investigated whether sequential transfer improves the solution of related TSPTW tasks compared with solving each task independently from scratch. Across three local-search baselines, LNS, VNS, and LKH-3, the results showed that the answer depends on the degree of structural continuity between consecutive tasks.

In the partial time-window expansion setting, where tasks are generated through progressive relaxation, the iterative protocol was consistently advantageous, especially on medium and large instances. It achieved higher success rates and more favorable penalized scores, showing that solutions from earlier tasks can be effectively reused when later tasks preserve the same underlying structure.

In the more difficult time-window swapping setting, the benefit of iteration became weaker and less uniform, as the perturbation of the route structure reduced the alignment between consecutive tasks.

\appendix

\section{Supplementary Tables}
\label{sec:supp-tables}

\begin{table*}[htbp]
    \tiny
    \centering
    \setlength{\tabcolsep}{2pt}
    \caption{Results for the Partial Time-Window Expansion Setting (VNS)}
    \label{tab:vns-expand}
    \begingroup
    \setlength{\arrayrulewidth}{0pt}
    \resizebox{0.80\textwidth}{!}{%
\begin{tabular}{ccrrrrr rrrrrc}
    \toprule
    \multirow{2}{*}{\textbf{Instance}} & \multirow{2}{*}{\textbf{Task}} & \multicolumn{5}{c}{\textbf{Iterative VNS}} & \multicolumn{5}{c}{\textbf{Standard VNS}} & \multirow{2}{*}{\textbf{Stat}}\\
    \cmidrule(lr){3-7}\cmidrule(lr){8-12}
     & & \textbf{mean} & \textbf{std} & \textbf{sr} & \textbf{succ. $\mu$} &  \textbf{succ. $\sigma$} & \textbf{mean} & \textbf{std} & \textbf{sr} & \textbf{succ. $\mu$} & \textbf{succ. $\sigma$} & \\ \midrule
     \multirow{5}{*}{20.10}
      & $T_1$ & 661.6 & 0 & 1 & 661.6 & 0 & 661.6 & 0 & 1 & 661.6 & 0 & (*) \\ & $T_2$ & 629.2 & 0 & 1 & 629.2 & 0 & 629.2 & 0 & 1 & 629.2 & 0 & (*) \\ & $T_3$ & 629.2 & 0 & 1 & 629.2 & 0 & 629.2 & 0 & 1 & 629.2 & 0 & (*) \\ & $T_4$ & 629.2 & 0 & 1 & 629.2 & 0 & 629.2 & 0 & 1 & 629.2 & 0 & (*) \\ & $T_5$ & 629.2 & 0 & 1 & 629.2 & 0 & 629.2 & 0 & 1 & 629.2 & 0 & (*) \\    \hline
     \multirow{5}{*}{20.20}
      & $T_1$ & 684.2 & 0 & 1 & 684.2 & 0 & 684.2 & 0 & 1 & 684.2 & 0 & (*) \\ & $T_2$ & 684.2 & 0 & 1 & 684.2 & 0 & 684.2 & 0 & 1 & 684.2 & 0 & (*) \\ & $T_3$ & 684.2 & 0 & 1 & 684.2 & 0 & 684.2 & 0 & 1 & 684.2 & 0 & (*) \\ & $T_4$ & 684.2 & 0 & 1 & 684.2 & 0 & 684.2 & 0 & 1 & 684.2 & 0 & (*) \\ & $T_5$ & 581.8 & 0 & 1 & 581.8 & 0 & 581.8 & 0 & 1 & 581.8 & 0 & (*) \\    \hline
    \multirow{5}{*}{40.10}
     & $T_1$ & 1100.6 & 0 & 1 & 1100.6 & 0 & 1100.6 & 0 & 1 & 1100.6 & 0 & (*) \\ & $T_2$ & 1084.3 & 0 & 1 & 1084.3 & 0 & 1085.93 & 4.97 & 1 & 1085.93 & 4.97 & (*) \\ & $T_3$ & 1083.62 & 0.13 & 1 & 1083.62 & 0.13 & 1.46e4 & 7.43e4 & 0.97 & 1085.31 & 4.72 & (+) \\ & $T_4$ & 1076.55 & 0.18 & 1 & 1076.55 & 0.18 & 3.98e4 & 1.18e5 & 0.9 & 1078.63 & 3.31 & (+) \\ & $T_5$ & 1043.82 & 0.13 & 1 & 1043.82 & 0.13 & 1.88e4 & 7.70e4 & 0.93 & 1045.58 & 3.43 & (+) \\    \hline
    \multirow{5}{*}{40.20}
     & $T_1$ & 1.60e5 & 8.71e5 & 0.97 & 1010.4 & 0 & 1.60e5 & 8.71e5 & 0.97 & 1010.4 & 0 & (*) \\ & $T_2$ & 984.09 & 3.02 & 1 & 984.09 & 3.02 & 1.95e5 & 5.97e5 & 0.9 & 984.97 & 5.61 & (*) \\ & $T_3$ & 983.6 & 2.17 & 1 & 983.6 & 2.17 & 3.62e5 & 7.34e5 & 0.8 & 984.58 & 3.74 & (+) \\ & $T_4$ & 956.9 & 2.17 & 1 & 956.9 & 2.17 & 1.90e5 & 5.49e5 & 0.87 & 959.54 & 6.53 & (+) \\ & $T_5$ & 956.9 & 2.17 & 1 & 956.9 & 2.17 & 1.42e5 & 5.06e5 & 0.9 & 958.28 & 6.89 & (+) \\    \hline
    \multirow{5}{*}{60.10}
     & $T_1$ & 9.99e5 & 2.06e6 & 0.6 & 601.17 & 5.52 & 9.99e5 & 2.06e6 & 0.6 & 601.17 & 5.52 & (*) \\ & $T_2$ & 1.55e5 & 8.47e5 & 0.97 & 516.41 & 13.44 & 1.66e4 & 6.26e4 & 0.93 & 524.07 & 15.06 & (-) \\ & $T_3$ & 510.73 & 5.81 & 1 & 510.73 & 5.81 & 527.37 & 21.48 & 1 & 527.37 & 21.48 & (+) \\ & $T_4$ & 477.43 & 5.51 & 1 & 477.43 & 5.51 & 9.39e4 & 4.94e5 & 0.93 & 493.96 & 21.43 & (+) \\ & $T_5$ & 473.13 & 2.24 & 1 & 473.13 & 2.24 & 500.07 & 26.09 & 1 & 500.07 & 26.09 & (+) \\    \hline
    \multirow{5}{*}{60.20}
     & $T_1$ & 6.51e7 & 5.32e7 & 0.03 & 1354 & 0 & 6.51e7 & 5.32e7 & 0.03 & 1354 & 0 & (*) \\ & $T_2$ & 1.11e7 & 1.51e7 & 0.43 & 1356.6 & 4.34 & 6.95e7 & 4.52e7 & 0.03 & 1357.4 & 0 & (+) \\ & $T_3$ & 7.24e6 & 1.06e7 & 0.6 & 1350.86 & 3.31 & 5.56e7 & 3.88e7 & 0.03 & 1355.2 & 0 & (+) \\ & $T_4$ & 2.66e6 & 5.00e6 & 0.7 & 1344.7 & 18.78 & 4.24e7 & 3.59e7 & 0.07 & 1266.05 & 4.31 & (+) \\ & $T_5$ & 9.88e5 & 2.72e6 & 0.87 & 1302.87 & 38.96 & 2.47e7 & 2.71e7 & 0.1 & 1211 & 9.35 & (+) \\    \hline
    \multirow{5}{*}{80.10}
     & $T_1$ & 1.79e7 & 1.20e7 & 0 & -- & -- & 1.79e7 & 1.20e7 & 0 & -- & -- & (*) \\ & $T_2$ & 2.64e6 & 3.98e6 & 0.43 & 643.31 & 15.07 & 1.90e7 & 1.25e7 & 0 & -- & -- & (+) \\ & $T_3$ & 8.40e5 & 2.57e6 & 0.77 & 623.57 & 13.13 & 2.11e7 & 1.95e7 & 0 & -- & -- & (+) \\ & $T_4$ & 4.41e5 & 2.09e6 & 0.93 & 598.89 & 11.69 & 7.94e6 & 1.34e7 & 0.2 & 633.33 & 15.1 & (+) \\ & $T_5$ & 4.41e5 & 2.09e6 & 0.93 & 594.57 & 10.54 & 5.19e6 & 8.66e6 & 0.33 & 639.7 & 22.86 & (+) \\    \hline
    \multirow{5}{*}{80.20}
     & $T_1$ & 5.60e7 & 3.44e7 & 0 & -- & -- & 5.60e7 & 3.44e7 & 0 & -- & -- & (*) \\ & $T_2$ & 9.26e6 & 1.33e7 & 0.07 & 695 & 12.73 & 3.13e7 & 2.31e7 & 0 & -- & -- & (+) \\ & $T_3$ & 1.40e6 & 2.95e6 & 0.53 & 681.75 & 13.06 & 2.60e7 & 1.91e7 & 0 & -- & -- & (+) \\ & $T_4$ & 2.10e5 & 6.16e5 & 0.83 & 650.16 & 15.72 & 1.39e7 & 1.55e7 & 0.13 & 667.75 & 10.14 & (+) \\ & $T_5$ & 623.37 & 12.63 & 1 & 623.37 & 12.63 & 9.60e6 & 1.09e7 & 0.13 & 683.75 & 20.76 & (+) \\    \hline
    \multirow{5}{*}{100.10}
     & $T_1$ & 4.43e8 & 1.48e8 & 0 & -- & -- & 4.43e8 & 1.48e8 & 0 & -- & -- & (*) \\ & $T_2$ & 1.38e8 & 8.20e7 & 0 & -- & -- & 3.62e8 & 1.35e8 & 0 & -- & -- & (+) \\ & $T_3$ & 2.66e7 & 3.11e7 & 0.1 & 786.33 & 33.71 & 2.71e8 & 1.26e8 & 0 & -- & -- & (+) \\ & $T_4$ & 4.68e6 & 7.11e6 & 0.3 & 745.67 & 23.78 & 2.78e8 & 1.24e8 & 0 & -- & -- & (+) \\ & $T_5$ & 3.59e5 & 8.78e5 & 0.77 & 738.22 & 21.06 & 2.65e8 & 1.09e8 & 0 & -- & -- & (+) \\    \hline
    \multirow{5}{*}{100.20}
     & $T_1$ & 1.22e8 & 5.84e7 & 0 & -- & -- & 1.22e8 & 5.84e7 & 0 & -- & -- & (*) \\ & $T_2$ & 7.37e6 & 1.14e7 & 0.17 & 708.6 & 14.05 & 1.10e8 & 6.48e7 & 0 & -- & -- & (+) \\ & $T_3$ & 3.76e5 & 9.09e5 & 0.77 & 693.13 & 20.56 & 9.73e7 & 7.08e7 & 0 & -- & -- & (+) \\ & $T_4$ & 8.40e4 & 3.24e5 & 0.93 & 678.18 & 19.7 & 8.26e7 & 7.03e7 & 0 & -- & -- & (+) \\ & $T_5$ & 3.40e4 & 1.83e5 & 0.97 & 667.38 & 16.7 & 5.13e7 & 3.73e7 & 0 & -- & -- & (+) \\    \hline
    \multirow{5}{*}{150.10}
     & $T_1$ & 6.24e9 & 9.00e8 & 0 & -- & -- & 6.24e9 & 9.00e8 & 0 & -- & -- & (*) \\ & $T_2$ & 2.79e9 & 7.51e8 & 0 & -- & -- & 6.14e9 & 1.59e9 & 0 & -- & -- & (+) \\ & $T_3$ & 1.22e9 & 4.66e8 & 0 & -- & -- & 4.56e9 & 1.10e9 & 0 & -- & -- & (+) \\ & $T_4$ & 5.44e8 & 3.12e8 & 0 & -- & -- & 5.16e9 & 1.17e9 & 0 & -- & -- & (+) \\ & $T_5$ & 3.20e8 & 2.34e8 & 0 & -- & -- & 4.37e9 & 9.91e8 & 0 & -- & -- & (+) \\    \hline
    \multirow{5}{*}{150.20}
     & $T_1$ & 5.09e9 & 8.73e8 & 0 & -- & -- & 5.09e9 & 8.73e8 & 0 & -- & -- & (*) \\ & $T_2$ & 2.03e9 & 6.44e8 & 0 & -- & -- & 4.59e9 & 1.11e9 & 0 & -- & -- & (+) \\ & $T_3$ & 1.02e9 & 4.38e8 & 0 & -- & -- & 4.12e9 & 8.23e8 & 0 & -- & -- & (+) \\ & $T_4$ & 5.02e8 & 2.53e8 & 0 & -- & -- & 3.41e9 & 1.14e9 & 0 & -- & -- & (+) \\ & $T_5$ & 2.67e8 & 1.78e8 & 0 & -- & -- & 3.29e9 & 9.34e8 & 0 & -- & -- & (+) \\    \hline
    \bottomrule
    \end{tabular}
    }
    \endgroup
\end{table*}

\begin{table*}[htbp]
    \tiny
    \centering
    \setlength{\tabcolsep}{2pt}
    \caption{Results for the Partial Time-Window Expansion Setting (LKH-3)}
    \label{tab:lkh-expand}
    \begingroup
    \setlength{\arrayrulewidth}{0pt}
    \resizebox{0.80\textwidth}{!}{%
\begin{tabular}{ccrrrrr rrrrrc}
    \toprule
\multirow{2}{*}{\textbf{Instance}} & \multirow{2}{*}{\textbf{Task}} & \multicolumn{5}{c}{\textbf{Iterative LKH-3}} & \multicolumn{5}{c}{\textbf{Standard LKH-3}} & \multirow{2}{*}{\textbf{Stat}}\\
 \cmidrule(lr){3-7}\cmidrule(lr){8-12}
 & & \textbf{mean} & \textbf{std} & \textbf{sr} & \textbf{succ. $\mu$} & \textbf{succ. $\sigma$} & \textbf{mean} & \textbf{std} & \textbf{sr} & \textbf{succ. $\mu$} & \textbf{succ. $\sigma$} & \\ \midrule
    \multirow{5}{*}{20.10}
     & $T_1$ & 661.6 & 0 & 1 & 661.6 & 0 & 661.6 & 0 & 1 & 661.6 & 0 & (*) \\ & $T_2$ & 629.2 & 0 & 1 & 629.2 & 0 & 629.2 & 0 & 1 & 629.2 & 0 & (*) \\ & $T_3$ & 629.2 & 0 & 1 & 629.2 & 0 & 1334.48 & 3862.97 & 0.97 & 629.2 & 0 & (*) \\ & $T_4$ & 629.2 & 0 & 1 & 629.2 & 0 & 637.43 & 9.58 & 1 & 637.43 & 9.58 & (+) \\ & $T_5$ & 629.2 & 0 & 1 & 629.2 & 0 & 634.27 & 8.55 & 1 & 634.27 & 8.55 & (+) \\    \hline
    \multirow{5}{*}{20.20}
     & $T_1$ & 684.2 & 0 & 1 & 684.2 & 0 & 684.2 & 0 & 1 & 684.2 & 0 & (*) \\ & $T_2$ & 684.2 & 0 & 1 & 684.2 & 0 & 684.2 & 0 & 1 & 684.2 & 0 & (*) \\ & $T_3$ & 684.2 & 0 & 1 & 684.2 & 0 & 684.2 & 0 & 1 & 684.2 & 0 & (*) \\ & $T_4$ & 684.2 & 0 & 1 & 684.2 & 0 & 684.2 & 0 & 1 & 684.2 & 0 & (*) \\ & $T_5$ & 581.8 & 0 & 1 & 581.8 & 0 & 581.8 & 0 & 1 & 581.8 & 0 & (*) \\    \hline
    \multirow{5}{*}{40.10}
     & $T_1$ & 1100.6 & 0 & 1 & 1100.6 & 0 & 1100.6 & 0 & 1 & 1100.6 & 0 & (*) \\ & $T_2$ & 1084.3 & 0 & 1 & 1084.3 & 0 & 1084.3 & 0 & 1 & 1084.3 & 0 & (*) \\ & $T_3$ & 1083.6 & 0 & 1 & 1083.6 & 0 & 1083.6 & 0 & 1 & 1083.6 & 0 & (*) \\ & $T_4$ & 1076.5 & 0 & 1 & 1076.5 & 0 & 1076.5 & 0 & 1 & 1076.5 & 0 & (*) \\ & $T_5$ & 1043.8 & 0 & 1 & 1043.8 & 0 & 1043.8 & 0 & 1 & 1043.8 & 0 & (*) \\    \hline
    \multirow{5}{*}{40.20}
     & $T_1$ & 1010.4 & 0 & 1 & 1010.4 & 0 & 1010.4 & 0 & 1 & 1010.4 & 0 & (*) \\ & $T_2$ & 983.3 & 0 & 1 & 983.3 & 0 & 983.3 & 0 & 1 & 983.3 & 0 & (*) \\ & $T_3$ & 983.2 & 0 & 1 & 983.2 & 0 & 983.2 & 0 & 1 & 983.2 & 0 & (*) \\ & $T_4$ & 983.2 & 0 & 1 & 983.2 & 0 & 972.07 & 13.63 & 1 & 972.07 & 13.63 & (*) \\ & $T_5$ & 983.2 & 0 & 1 & 983.2 & 0 & 3.73e5 & 5.35e5 & 0.67 & 990.96 & 14.54 & (+) \\    \hline
    \multirow{5}{*}{60.10}
     & $T_1$ & 2.89e6 & 6.71e6 & 0.83 & 591 & 0 & 2.89e6 & 6.71e6 & 0.83 & 591 & 0 & (*) \\ & $T_2$ & 504.7 & 1.82 & 1 & 504.7 & 1.82 & 9.02e5 & 3.66e6 & 0.03 & 509 & 0 & (+) \\ & $T_3$ & 503.67 & 1.73 & 1 & 503.67 & 1.73 & 4.57e6 & 1.90e7 & 0.13 & 506 & 4.24 & (+) \\ & $T_4$ & 471 & 0 & 1 & 471 & 0 & 6.37e7 & 6.44e7 & 0 & -- & -- & (+) \\ & $T_5$ & 469.4 & 0.81 & 1 & 469.4 & 0.81 & 3.60e7 & 3.40e7 & 0.03 & 490 & 0 & (+) \\    \hline
    \multirow{5}{*}{60.20}
     & $T_1$ & 1353.5 & 0 & 1 & 1353.5 & 0 & 1353.5 & 0 & 1 & 1353.5 & 0 & (*) \\ & $T_2$ & 1353.5 & 0 & 1 & 1353.5 & 0 & 1353.5 & 0 & 1 & 1353.5 & 0 & (*) \\ & $T_3$ & 1349.1 & 0 & 1 & 1349.1 & 0 & 1349.1 & 0 & 1 & 1349.1 & 0 & (*) \\ & $T_4$ & 1348 & 0 & 1 & 1348 & 0 & 1263 & 0 & 1 & 1263 & 0 & (-) \\ & $T_5$ & 1290.6 & 0 & 1 & 1290.6 & 0 & 3.68e7 & 1.03e7 & 0 & -- & -- & (+) \\    \hline
    \multirow{5}{*}{80.10}
     & $T_1$ & 9.29e6 & 9.22e6 & 0 & -- & -- & 9.29e6 & 9.22e6 & 0 & -- & -- & (*) \\ & $T_2$ & 2.26e6 & 6.45e6 & 0.77 & 618.26 & 2.68 & 3.77e7 & 5.19e7 & 0.2 & 626.5 & 8.5 & (+) \\ & $T_3$ & 2.16e6 & 6.47e6 & 0.87 & 612.92 & 7.58 & 1.39e8 & 9.23e7 & 0 & -- & -- & (+) \\ & $T_4$ & 5.83e5 & 2.19e6 & 0.9 & 593.59 & 12.22 & 2.41e8 & 1.84e8 & 0 & -- & -- & (+) \\ & $T_5$ & 5.83e5 & 2.19e6 & 0.9 & 589.04 & 12.48 & 1.66e8 & 1.02e8 & 0 & -- & -- & (+) \\    \hline
    \multirow{5}{*}{80.20}
     & $T_1$ & 2.52e6 & 3.97e6 & 0.03 & 674 & 0 & 2.52e6 & 3.97e6 & 0.03 & 674 & 0 & (*) \\ & $T_2$ & 1.45e6 & 3.77e6 & 0.87 & 666 & 0 & 3.32e6 & 4.98e6 & 0.67 & 666 & 0 & (*) \\ & $T_3$ & 1.45e6 & 3.77e6 & 0.87 & 664.15 & 4.42 & 3.23e8 & 2.29e8 & 0 & -- & -- & (+) \\ & $T_4$ & 1.20e6 & 3.19e6 & 0.87 & 661.96 & 8.24 & 5.76e8 & 2.37e8 & 0 & -- & -- & (+) \\ & $T_5$ & 8.15e5 & 2.53e6 & 0.9 & 650.96 & 14.78 & 4.50e8 & 1.96e8 & 0 & -- & -- & (+) \\    \hline
    \multirow{5}{*}{100.10}
     & $T_1$ & 2.34e7 & 2.51e7 & 0.43 & 770 & 0 & 2.34e7 & 2.51e7 & 0.43 & 770 & 0 & (*) \\ & $T_2$ & 6.74e4 & 3.21e5 & 0.93 & 759.04 & 0.19 & 2.51e7 & 2.57e7 & 0.27 & 759.75 & 1.16 & (+) \\ & $T_3$ & 9058.63 & 4.57e4 & 0.97 & 721.86 & 23.38 & 8.59e8 & 6.70e8 & 0 & -- & -- & (+) \\ & $T_4$ & 715.57 & 25.29 & 1 & 715.57 & 25.29 & 1.71e9 & 8.68e8 & 0 & -- & -- & (+) \\ & $T_5$ & 711.13 & 21.2 & 1 & 711.13 & 21.2 & 3.43e9 & 7.36e8 & 0 & -- & -- & (+) \\    \hline
    \multirow{5}{*}{100.20}
     & $T_1$ & 8.51e7 & 4.10e7 & 0.07 & 658 & 1.41 & 8.51e7 & 4.10e7 & 0.07 & 658 & 1.41 & (*) \\ & $T_2$ & 6.08e7 & 3.21e7 & 0.13 & 643.5 & 3.11 & 1.95e8 & 1.13e8 & 0 & -- & -- & (+) \\ & $T_3$ & 5.67e7 & 3.24e7 & 0.13 & 625.5 & 2.38 & 1.69e8 & 1.10e8 & 0 & -- & -- & (+) \\ & $T_4$ & 2.58e7 & 1.59e7 & 0.13 & 625 & 2.45 & 1.42e8 & 6.08e7 & 0 & -- & -- & (+) \\ & $T_5$ & 2.51e7 & 1.46e7 & 0.13 & 621.75 & 3.86 & 4.84e8 & 2.66e8 & 0 & -- & -- & (+) \\    \hline
    \multirow{5}{*}{150.10}
     & $T_1$ & 1.18e9 & 2.31e9 & 0 & -- & -- & 1.18e9 & 2.31e9 & 0 & -- & -- & (*) \\ & $T_2$ & 3.90e8 & 5.17e8 & 0 & -- & -- & 2.28e9 & 3.35e9 & 0 & -- & -- & (+) \\ & $T_3$ & 2.71e8 & 4.22e8 & 0.3 & 936.33 & 7.4 & 7.43e9 & 7.90e9 & 0 & -- & -- & (+) \\ & $T_4$ & 2.23e8 & 3.75e8 & 0.37 & 932.45 & 8.25 & 1.11e10 & 8.11e9 & 0 & -- & -- & (+) \\ & $T_5$ & 1.63e8 & 2.93e8 & 0.37 & 930.82 & 8.95 & 1.03e10 & 4.13e9 & 0 & -- & -- & (+) \\    \hline
    \multirow{5}{*}{150.20}
     & $T_1$ & 1.56e9 & 2.72e9 & 0 & -- & -- & 1.56e9 & 2.72e9 & 0 & -- & -- & (*) \\ & $T_2$ & 7.72e8 & 8.29e8 & 0 & -- & -- & 3.47e9 & 1.90e9 & 0 & -- & -- & (+) \\ & $T_3$ & 5.53e8 & 6.37e8 & 0.03 & 739 & 0 & 6.16e9 & 8.82e9 & 0 & -- & -- & (+) \\ & $T_4$ & 4.73e8 & 5.38e8 & 0.03 & 707 & 0 & 4.43e9 & 2.10e9 & 0 & -- & -- & (+) \\ & $T_5$ & 3.08e8 & 3.43e8 & 0.03 & 647 & 0 & 5.12e9 & 3.87e9 & 0 & -- & -- & (+) \\    \hline
    \bottomrule
    \end{tabular}
    }
    \endgroup
\end{table*}

\begin{table*}[htbp]
    \tiny
    \centering
    \setlength{\tabcolsep}{2pt}
    \caption{Results for the Swap-Additive Time-Window Setting (VNS)}
    \label{tab:vns-swap}
    \begingroup
    \setlength{\arrayrulewidth}{0pt}
    \resizebox{0.80\textwidth}{!}{%
\begin{tabular}{ccrrrrr rrrrrc}
    \toprule
    \multirow{2}{*}{\textbf{Instance}} & \multirow{2}{*}{\textbf{Task}} & \multicolumn{5}{c}{\textbf{Iterative VNS}} & \multicolumn{5}{c}{\textbf{Standard VNS}} & \multirow{2}{*}{\textbf{Stat}}\\
    \cmidrule(lr){3-7}\cmidrule(lr){8-12}
     & & \textbf{mean} & \textbf{std} & \textbf{sr} & \textbf{succ. $\mu$} &  \textbf{succ. $\sigma$} & \textbf{mean} & \textbf{std} & \textbf{sr} & \textbf{succ. $\mu$} & \textbf{succ. $\sigma$} & \\ \midrule
    \multirow{5}{*}{20.10}
     & $T_1$ & 713.2 & 0 & 1 & 713.20 & 0 & 713.2 & 0 & 1 & 713.20 & 0 & (*) \\ & $T_2$ & 706.3 & 0 & 1 & 706.30 & 0 & 706.3 & 0 & 1 & 706.30 & 0 & (*) \\ & $T_3$ & 798.4 & 0 & 1 & 798.40 & 0 & 798.4 & 0 & 1 & 798.40 & 0 & (*) \\ & $T_4$ & 919.5 & 0 & 1 & 919.50 & 0 & 919.5 & 0 & 1 & 919.50 & 0 & (*) \\ & $T_5$ & 934.2 & 0 & 1 & 934.20 & 0 & 934.2 & 0 & 1 & 934.20 & 0 & (*) \\    \hline
    \multirow{5}{*}{20.20}
     & $T_1$ & 781 & 0 & 1 & 781 & 0 & 781 & 0 & 1 & 781 & 0 & (*) \\ & $T_2$ & 831 & 0 & 1 & 831 & 0 & 831 & 0 & 1 & 831 & 0 & (*) \\ & $T_3$ & 842.4 & 0 & 1 & 842.40 & 0 & 842.4 & 0 & 1 & 842.40 & 0 & (*) \\ & $T_4$ & 856.2 & 0 & 1 & 856.20 & 0 & 856.2 & 0 & 1 & 856.20 & 0 & (*) \\ & $T_5$ & 894.9 & 0 & 1 & 894.90 & 0 & 894.9 & 0 & 1 & 894.90 & 0 & (*) \\    \hline
    \multirow{5}{*}{40.10}
     & $T_1$ & 1429.1 & 0.55 & 1 & 1429.10 & 0.55 & 1429.1 & 0.55 & 1 & 1429.10 & 0.55 & (*) \\ & $T_2$ & 1523.16 & 2.62 & 1 & 1523.16 & 2.62 & 1524.61 & 3.1 & 1 & 1524.61 & 3.10 & (*) \\ & $T_3$ & 1493.49 & 3.33 & 1 & 1493.49 & 3.33 & 1492.54 & 3.12 & 1 & 1492.54 & 3.12 & (*) \\ & $T_4$ & 1596.35 & 2.87 & 1 & 1596.35 & 2.87 & 1596.67 & 3.16 & 1 & 1596.67 & 3.16 & (*) \\ & $T_5$ & 1740.54 & 5.29 & 1 & 1740.54 & 5.29 & 1741.01 & 4.47 & 1 & 1741.01 & 4.47 & (*) \\    \hline
    \multirow{5}{*}{40.20}
     & $T_1$ & 1320.75 & 5.68 & 1 & 1320.75 & 5.68 & 1320.75 & 5.68 & 1 & 1320.75 & 5.68 & (*) \\ & $T_2$ & 1361.32 & 2.38 & 1 & 1361.32 & 2.38 & 1364.08 & 10.98 & 1 & 1364.08 & 10.98 & (*) \\ & $T_3$ & 1420 & 3.48 & 1 & 1420 & 3.48 & 1421.18 & 7.26 & 1 & 1421.18 & 7.26 & (*) \\ & $T_4$ & 1438.9 & 0 & 1 & 1438.90 & 0 & 1440.17 & 4.84 & 1 & 1440.17 & 4.84 & (*) \\ & $T_5$ & 1513.47 & 7.14 & 1 & 1513.47 & 7.14 & 1513.89 & 7.74 & 1 & 1513.89 & 7.74 & (*) \\    \hline
    \multirow{5}{*}{60.10}
     & $T_1$ & 5.06e6 & 1.01e7 & 0.53 & 1617.96 & 17.90 & 5.06e6 & 1.01e7 & 0.53 & 1617.96 & 17.90 & (*) \\ & $T_2$ & 3.61e6 & 9.85e6 & 0.73 & 1660.33 & 22.78 & 1.5e6 & 6.16e6 & 0.8 & 1660.04 & 21.12 & (*) \\ & $T_3$ & 1639.13 & 15.14 & 1 & 1639.13 & 15.14 & 2.2e6 & 6.54e6 & 0.7 & 1671.13 & 30.31 & (+) \\ & $T_4$ & 1726.2 & 13.73 & 1 & 1726.20 & 13.73 & 5.97e6 & 1.33e7 & 0.67 & 1767.93 & 37.01 & (+) \\ & $T_5$ & 1.15e6 & 5.22e6 & 0.7 & 1834.12 & 26.34 & 2.54e6 & 6.45e6 & 0.57 & 1845.69 & 27.53 & (+) \\    \hline
    \multirow{5}{*}{60.20}
     & $T_1$ & 3.58e5 & 1.05e6 & 0.73 & 707.64 & 16.37 & 3.58e5 & 1.05e6 & 0.73 & 707.64 & 16.37 & (*) \\ & $T_2$ & 1.23e5 & 6.7e5 & 0.97 & 717.93 & 13.80 & 1.17e5 & 6.35e5 & 0.97 & 725.97 & 18.23 & (*) \\ & $T_3$ & 751.63 & 12.52 & 1 & 751.63 & 12.52 & 1.49e5 & 7.1e5 & 0.93 & 754.39 & 14.66 & (*) \\ & $T_4$ & 737.6 & 5.57 & 1 & 737.60 & 5.57 & 751.83 & 15 & 1 & 751.83 & 15 & (+) \\ & $T_5$ & 752.1 & 14.6 & 1 & 752.10 & 14.60 & 3.55e5 & 1.08e6 & 0.9 & 753 & 12.75 & (*) \\    \hline
    \multirow{5}{*}{80.10}
     & $T_1$ & 9.45e6 & 9.13e6 & 0.2 & 937 & 16.63 & 9.45e6 & 9.13e6 & 0.2 & 937 & 16.63 & (*) \\ & $T_2$ & 3.26e5 & 1.18e6 & 0.87 & 898.62 & 16.20 & 4.05e6 & 6.03e6 & 0.43 & 927.92 & 23.45 & (+) \\ & $T_3$ & 3.66e6 & 6.03e6 & 0.43 & 924.54 & 15.83 & 4.96e6 & 8.64e6 & 0.47 & 925.93 & 28.16 & (*) \\ & $T_4$ & 4.94e5 & 2.7e6 & 0.97 & 883.41 & 18.34 & 4.02e6 & 6.42e6 & 0.47 & 919.71 & 27.93 & (+) \\      & $T_5$ & 4.65e6 & 7.4e6 & 0.47 & 916.86 & 19.49 & 6.03e6 & 6.87e6 & 0.23 & 930 & 17.73 & (*) \\    \hline
    \multirow{5}{*}{80.20}
     & $T_1$ & 9.5e6 & 9.1e6 & 0.07 & 1023.50 & 27.58 & 9.5e6 & 9.1e6 & 0.07 & 1023.50 & 27.58 & (*) \\ & $T_2$ & 6.72e6 & 7.25e6 & 0.23 & 1076.57 & 27.02 & 7.96e6 & 7.28e6 & 0.17 & 1065 & 23 & (*) \\ & $T_3$ & 7.91e6 & 7.03e6 & 0.23 & 1077.29 & 40.65 & 6.71e6 & 5.74e6 & 0.23 & 1082.29 & 25.83 & (*) \\ & $T_4$ & 3.95e6 & 5.38e6 & 0.53 & 1087.25 & 26.46 & 6.48e6 & 5.92e6 & 0.23 & 1107.57 & 24.26 & (+) \\ & $T_5$ & 4.16e6 & 5.07e6 & 0.27 & 1170.75 & 32.81 & 4.47e6 & 4.99e6 & 0.2 & 1197.17 & 27.90 & (*) \\    \hline
    \multirow{5}{*}{100.10}
     & $T_1$ & 1.07e8 & 5.03e7 & 0 & -- & -- & 1.07e8 & 5.03e7 & 0 & -- & -- & (*) \\ & $T_2$ & 6.34e7 & 2.94e7 & 0 & -- & -- & 9.29e7 & 4.12e7 & 0 & -- & -- & (+) \\ & $T_3$ & 1.93e7 & 1.21e7 & 0.03 & 1332 & 0 & 9.14e7 & 4.49e7 & 0 & -- & -- & (+) \\ & $T_4$ & 4.83e7 & 2.3e7 & 0 & -- & -- & 8.33e7 & 4.15e7 & 0 & -- & -- & (+) \\ & $T_5$ & 6.66e7 & 5.11e7 & 0 & -- & -- & 7.94e7 & 5.19e7 & 0 & -- & -- & (*) \\    \hline
    \multirow{5}{*}{100.20}
     & $T_1$ & 2.96e7 & 1.96e7 & 0 & -- & -- & 2.96e7 & 1.96e7 & 0 & -- & -- & (*) \\ & $T_2$ & 7.9e6 & 8.27e6 & 0.27 & 1185.88 & 41.25 & 2.5e7 & 1.33e7 & 0 & -- & -- & (+) \\ & $T_3$ & 1.85e7 & 1.2e7 & 0.03 & 1275 & 0 & 1.9e7 & 1.67e7 & 0 & -- & -- & (*) \\      & $T_4$ & 1.24e7 & 1.13e7 & 0 & -- & -- & 2.01e7 & 1.65e7 & 0.03 & 1414 & 0 & (*) \\ & $T_5$ & 1.42e7 & 1.36e7 & 0.1 & 1402 & 42.58 & 1.92e7 & 1.42e7 & 0 & -- & -- & (*) \\    \hline
    \multirow{5}{*}{150.10}
     & $T_1$ & 1.17e9 & 3.22e8 & 0 & -- & -- & 1.17e9 & 3.22e8 & 0 & -- & -- & (*) \\ & $T_2$ & 4.23e8 & 1.74e8 & 0 & -- & -- & 1.14e9 & 3.68e8 & 0 & -- & -- & (+) \\ & $T_3$ & 7.6e8 & 3.02e8 & 0 & -- & -- & 1.14e9 & 2.78e8 & 0 & -- & -- & (+) \\ & $T_4$ & 8.3e8 & 3.79e8 & 0 & -- & -- & 1.01e9 & 3.72e8 & 0 & -- & -- & (+) \\ & $T_5$ & 6.11e8 & 2.79e8 & 0 & -- & -- & 1.05e9 & 3.58e8 & 0 & -- & -- & (+) \\    \hline
    \multirow{5}{*}{150.20}
     & $T_1$ & 8.36e8 & 3.49e8 & 0 & -- & -- & 8.36e8 & 3.49e8 & 0 & -- & -- & (*) \\ & $T_2$ & 1.92e8 & 1.55e8 & 0 & -- & -- & 7.81e8 & 3.48e8 & 0 & -- & -- & (+) \\ & $T_3$ & 8.53e7 & 8.48e7 & 0 & -- & -- & 7.97e8 & 3.49e8 & 0 & -- & -- & (+) \\ & $T_4$ & 7.21e7 & 8.71e7 & 0.03 & 1609 & 0 & 7.14e8 & 3.33e8 & 0 & -- & -- & (+) \\ & $T_5$ & 5.67e8 & 2.95e8 & 0 & -- & -- & 8.01e8 & 3.27e8 & 0 & -- & -- & (+) \\    \hline
    \bottomrule
    \end{tabular}
    }
    \endgroup
\end{table*}

\begin{table*}[htbp]
    \tiny
    \centering
    \setlength{\tabcolsep}{2pt}
    \caption{Results for the Swap-Additive Time-Window Setting (LKH-3)}
    \label{tab:lkh-swap}
    \begingroup
    \setlength{\arrayrulewidth}{0pt}
    \resizebox{0.80\textwidth}{!}{%
\begin{tabular}{ccrrrrr rrrrrc}
    \toprule
\multirow{2}{*}{\textbf{Instance}} & \multirow{2}{*}{\textbf{Task}} & \multicolumn{5}{c}{\textbf{Iterative LKH-3}} & \multicolumn{5}{c}{\textbf{Standard LKH-3}} & \multirow{2}{*}{\textbf{Stat}}\\
 \cmidrule(lr){3-7}\cmidrule(lr){8-12}
 & & \textbf{mean} & \textbf{std} & \textbf{sr} & \textbf{succ. $\mu$} & \textbf{succ. $\sigma$} & \textbf{mean} & \textbf{std} & \textbf{sr} & \textbf{succ. $\mu$} & \textbf{succ. $\sigma$} & \\ \midrule
    \multirow{5}{*}{20.10}
     & $T_1$ & 713.20 & 0 & 1 & 713.20 & 0 & 713.20 & 0 & 1 & 713.20 & 0 & (*) \\ & $T_2$ & 706.30 & 0 & 1 & 706.30 & 0 & 706.30 & 0 & 1 & 706.30 & 0 & (*) \\ & $T_3$ & 798.40 & 0 & 1 & 798.40 & 0 & 798.40 & 0 & 1 & 798.40 & 0 & (*) \\ & $T_4$ & 919.50 & 0 & 1 & 919.50 & 0 & 919.50 & 0 & 1 & 919.50 & 0 & (*) \\ & $T_5$ & 934.20 & 0 & 1 & 934.20 & 0 & 934.20 & 0 & 1 & 934.20 & 0 & (*) \\    \hline
    \multirow{5}{*}{20.20}
     & $T_1$ & 781 & 0 & 1 & 781 & 0 & 781 & 0 & 1 & 781 & 0 & (*) \\ & $T_2$ & 1.34e5 & 0 & 0 & -- & -- & 1.34e5 & 0 & 0 & -- & -- & (*) \\ & $T_3$ & 1.42e5 & 0 & 0 & -- & -- & 1.42e5 & 0 & 0 & -- & -- & (*) \\ & $T_4$ & 1.46e5 & 0 & 0 & -- & -- & 1.46e5 & 0 & 0 & -- & -- & (*) \\      & $T_5$ & 3.21e4 & 3.17e4 & 0.50 & 894.90 & 0 & 5.92e4 & 1.58e4 & 0.07 & 894.90 & 0 & (+) \\    \hline
    \multirow{5}{*}{40.10}
     & $T_1$ & 1437.50 & 0 & 1 & 1437.50 & 0 & 1437.50 & 0 & 1 & 1437.50 & 0 & (*) \\ & $T_2$ & 1521.80 & 0 & 1 & 1521.80 & 0 & 1521.80 & 0 & 1 & 1521.80 & 0 & (*) \\ & $T_3$ & 1490 & 0 & 1 & 1490 & 0 & 1490 & 0 & 1 & 1490 & 0 & (*) \\ & $T_4$ & 1593.20 & 0 & 1 & 1593.20 & 0 & 1593.20 & 0 & 1 & 1593.20 & 0 & (*) \\ & $T_5$ & 1734.30 & 0 & 1 & 1734.30 & 0 & 1734.30 & 0 & 1 & 1734.30 & 0 & (*) \\    \hline
    \multirow{5}{*}{40.20}
     & $T_1$ & 1357.83 & 10.75 & 1 & 1357.83 & 10.75 & 1357.83 & 10.75 & 1 & 1357.83 & 10.75 & (*) \\ & $T_2$ & 1364.19 & 7.29 & 1 & 1364.19 & 7.29 & 1376.94 & 11.71 & 1 & 1376.94 & 11.71 & (+) \\ & $T_3$ & 1432.55 & 11.47 & 1 & 1432.55 & 11.47 & 1441.59 & 5.73 & 1 & 1441.59 & 5.73 & (+) \\ & $T_4$ & 1452.25 & 11.47 & 1 & 1452.25 & 11.47 & 1458.32 & 9.22 & 1 & 1458.32 & 9.22 & (+) \\ & $T_5$ & 1511.80 & 0 & 1 & 1511.80 & 0 & 1511.80 & 0 & 1 & 1511.80 & 0 & (*) \\    \hline
    \multirow{5}{*}{60.10}
     & $T_1$ & 3.12e8 & 1.68e8 & 0.20 & 1589.30 & 0 & 3.12e8 & 1.68e8 & 0.20 & 1589.30 & 0 & (*) \\ & $T_2$ & 3.86e7 & 9.57e7 & 0.80 & 1632.21 & 4.17 & 3e7 & 6.2e7 & 0.77 & 1632.10 & 4.24 & (*) \\ & $T_3$ & 1626.85 & 0.18 & 1 & 1626.85 & 0.18 & 4.2e7 & 9.89e7 & 0.77 & 1630.10 & 4.78 & (+) \\ & $T_4$ & 1714.88 & 6.27 & 1 & 1714.88 & 6.27 & 3.5e7 & 4.69e7 & 0.40 & 1723.35 & 6.96 & (+) \\ & $T_5$ & 1815.17 & 3.91 & 1 & 1815.17 & 3.91 & 2.74e6 & 1.04e7 & 0.93 & 1816.22 & 0.13 & (+) \\    \hline
    \multirow{5}{*}{60.20}
     & $T_1$ & 1.79e7 & 3.9e7 & 0.27 & 675.62 & 4.60 & 1.79e7 & 3.9e7 & 0.27 & 675.62 & 4.60 & (*) \\ & $T_2$ & 5.49e6 & 2.11e7 & 0.93 & 695 & 0 & 695 & 0 & 1 & 695 & 0 & (*) \\ & $T_3$ & 729.37 & 2.01 & 1 & 729.37 & 2.01 & 4.06e7 & 9.33e7 & 0.83 & 734.72 & 5.61 & (+) \\ & $T_4$ & 729.37 & 2.01 & 1 & 729.37 & 2.01 & 4.06e7 & 9.33e7 & 0.83 & 734.72 & 5.61 & (+) \\ & $T_5$ & 729 & 0 & 1 & 729 & 0 & 9.86e6 & 3.67e7 & 0.90 & 729 & 0 & (*) \\    \hline
    \multirow{5}{*}{80.10}
     & $T_1$ & 9.05e8 & 2.19e8 & 0 & -- & -- & 9.05e8 & 2.19e8 & 0 & -- & -- & (*) \\ & $T_2$ & 8.02e8 & 2.75e8 & 0 & -- & -- & 9.76e8 & 2.39e8 & 0 & -- & -- & (+) \\ & $T_3$ & 6.87e8 & 3.64e8 & 0 & -- & -- & 9.41e8 & 3.05e8 & 0 & -- & -- & (+) \\ & $T_4$ & 5.31e8 & 4.14e8 & 0 & -- & -- & 6.5e8 & 3.49e8 & 0 & -- & -- & (*) \\ & $T_5$ & 5.07e8 & 2.75e8 & 0 & -- & -- & 7.23e8 & 1.95e8 & 0 & -- & -- & (+) \\    \hline
    \multirow{5}{*}{80.20}
     & $T_1$ & 5.89e8 & 1.54e8 & 0 & -- & -- & 5.89e8 & 1.54e8 & 0 & -- & -- & (*) \\ & $T_2$ & 6.25e8 & 1.03e8 & 0 & -- & -- & 6.68e8 & 1.32e8 & 0 & -- & -- & (*) \\ & $T_3$ & 6.33e8 & 6.44e7 & 0 & -- & -- & 6.96e8 & 1.46e8 & 0 & -- & -- & (+) \\ & $T_4$ & 6.31e8 & 7.51e7 & 0 & -- & -- & 8.61e8 & 3.76e8 & 0 & -- & -- & (+) \\ & $T_5$ & 7.71e8 & 1.04e8 & 0 & -- & -- & 8.38e8 & 1.26e8 & 0 & -- & -- & (+) \\    \hline
    \multirow{5}{*}{100.10}
     & $T_1$ & 1.96e9 & 7.75e8 & 0 & -- & -- & 1.96e9 & 7.75e8 & 0 & -- & -- & (*) \\ & $T_2$ & 2.07e9 & 7.08e8 & 0 & -- & -- & 2.37e9 & 7.3e8 & 0 & -- & -- & (*) \\ & $T_3$ & 1.73e9 & 5.5e8 & 0 & -- & -- & 2.19e9 & 6.67e8 & 0 & -- & -- & (+) \\ & $T_4$ & 2.1e9 & 8e8 & 0 & -- & -- & 3.1e9 & 1.22e9 & 0 & -- & -- & (+) \\ & $T_5$ & 2.29e9 & 1.01e9 & 0 & -- & -- & 3.85e9 & 1.13e9 & 0 & -- & -- & (+) \\    \hline
    \multirow{5}{*}{100.20}
     & $T_1$ & 1.73e9 & 1.33e9 & 0 & -- & -- & 1.73e9 & 1.33e9 & 0 & -- & -- & (*) \\ & $T_2$ & 1.38e9 & 1.3e9 & 0 & -- & -- & 2.69e9 & 1.31e9 & 0 & -- & -- & (+) \\      & $T_3$ & 1.38e9 & 1.23e9 & 0 & -- & -- & 1.94e9 & 1.35e9 & 0 & -- & -- & (*) \\ & $T_4$ & 1.62e9 & 1.11e9 & 0 & -- & -- & 1.48e9 & 9.55e8 & 0 & -- & -- & (*) \\ & $T_5$ & 8.69e8 & 9.32e8 & 0 & -- & -- & 7.42e8 & 5.73e8 & 0 & -- & -- & (*) \\    \hline
    \multirow{5}{*}{150.10}
     & $T_1$ & 2.18e10 & 1.41e10 & 0 & -- & -- & 2.18e10 & 1.41e10 & 0 & -- & -- & (*) \\ & $T_2$ & 1.99e10 & 9.91e9 & 0 & -- & -- & 2.24e10 & 1.27e10 & 0 & -- & -- & (*) \\ & $T_3$ & 1.77e10 & 1.18e10 & 0 & -- & -- & 1.8e10 & 7.77e9 & 0 & -- & -- & (*) \\ & $T_4$ & 1.86e10 & 1.07e10 & 0 & -- & -- & 1.87e10 & 8.45e9 & 0 & -- & -- & (*) \\ & $T_5$ & 1.32e10 & 6.33e9 & 0 & -- & -- & 2.01e10 & 9.1e9 & 0 & -- & -- & (+) \\    \hline
    \multirow{5}{*}{150.20}
     & $T_1$ & 1.61e10 & 5.46e9 & 0 & -- & -- & 1.61e10 & 5.46e9 & 0 & -- & -- & (*) \\ & $T_2$ & 1.82e10 & 7.34e9 & 0 & -- & -- & 1.67e10 & 9.81e9 & 0 & -- & -- & (*) \\ & $T_3$ & 1.44e10 & 5.56e9 & 0 & -- & -- & 1.4e10 & 7.16e9 & 0 & -- & -- & (*) \\ & $T_4$ & 1.09e10 & 3.63e9 & 0 & -- & -- & 1.34e10 & 8.91e9 & 0 & -- & -- & (*) \\ & $T_5$ & 1.35e10 & 6.58e9 & 0 & -- & -- & 1.37e10 & 6.46e9 & 0 & -- & -- & (*) \\    \hline
    \bottomrule
    \end{tabular}
    }
    \endgroup
\end{table*}

\clearpage
\bibliographystyle{unsrt}
\bibliography{references}

\end{document}